\documentclass[10pt,twocolumn,letterpaper]{article}

\usepackage{cvpr}
\usepackage{times}
\usepackage{epsfig}
\usepackage{graphicx}
\usepackage{amsmath}
\usepackage{amssymb}
\usepackage{booktabs}
\usepackage{enumerate}
\usepackage{multirow}

\usepackage[pagebackref=true,breaklinks=true,letterpaper=true,colorlinks,bookmarks=false]{hyperref}

\cvprfinalcopy 

\def\cvprPaperID{4244} 

\ifcvprfinal\fi
\begin{document}

\title{Diversified Arbitrary Style Transfer via Deep Feature Perturbation}

\author{Zhizhong Wang,\hspace{0.4cm}Lei Zhao$^*$,\hspace{0.4cm}Haibo Chen,\hspace{0.4cm}Lihong Qiu,\\
	Qihang Mo,\hspace{0.4cm}Sihuan Lin,\hspace{0.4cm}Wei Xing,\hspace{0.4cm}Dongming Lu\\
	\\
	College of Computer Science and Technology, Zhejiang University\\
	{\tt\small \{endywon, cszhl, feng123, zjusheldon, moqihang, linsh, wxing, ldm\}@zju.edu.cn}
}

\maketitle
\thispagestyle{empty}
\renewcommand{\thefootnote}{\fnsymbol{footnote}}
\setcounter{footnote}{-1}
\begin{abstract}
Image style transfer is an underdetermined problem\footnote{* Corresponding author}, where a large number of solutions can satisfy the same constraint (the content and style). Although there have been some efforts to improve the diversity of style transfer by introducing an alternative diversity loss, they have restricted generalization, limited diversity and poor scalability. In this paper, we tackle these limitations and propose a simple yet effective method for diversified arbitrary style transfer. The key idea of our method is an operation called deep feature perturbation (DFP), which uses an orthogonal random noise matrix to perturb the deep image feature maps while keeping the original style information unchanged. Our DFP operation can be easily integrated into many existing WCT (whitening and coloring transform)-based methods, and empower them to generate diverse results for arbitrary styles. Experimental results demonstrate that this learning-free and universal method can greatly increase the diversity while maintaining the quality of stylization.

\end{abstract}

\section{Introduction}
\label{introduction}
Style transfer, or to repaint an existing image with the style of another, is considered as a challenging but interesting task in both academia and industry. Recently, the pioneering works of Gatys \etal~\cite{gatys2015neural,gatys2015texture,gatys2016image} have proved that the correlations (\ie, Gram matrix) between feature maps extracted from a pre-trained deep convolutional neural network (DCNN) can represent the style of an image well. Since then, significant efforts have been made to improve in many aspects including efficiency~\cite{ulyanov2016texture,johnson2016perceptual,li2016precomputed}, quality~\cite{li2016combining,wang2017multimodal,liao2017visual,gu2018arbitrary}, generality~\cite{chen2017stylebank,dumoulin2017learned,huang2017arbitrary,li2017universal,sheng2018avatar,lu2019closed}, user control~\cite{champandard2016semantic,gatys2017controlling} and photorealism~\cite{luan2017deep,li2018closed,yoo2019photorealistic}, etc. However, despite the remarkable success, these methods often neglect an important aspect, \ie, the diversity, since many of the applications (\eg, art creation and creative design) are required to satisfy the preferences of different users.

In terms of diversity, one common explanation is that, image style transfer is an underdetermined problem, where a large number of solutions can satisfy the same content and style, just like the results generated by different methods could all be visually pleasing and perceptually correct. However, the lack of meaningful variations in vanilla style transfer mechanism~\cite{gatys2016image,johnson2016perceptual,ulyanov2016texture} hampers the emergence of diversity, as the optimization-based methods often converge to the similar local optimum, while the feed-forward networks only produce fixed outputs for the fixed inputs. 

Although challenging and meaningful, unfortunately, this problem has barely received enough attention and there are only a few efforts to solve it. For instance, based on the feed-forward networks, Li \etal~\cite{li2017diversified} introduced a diversity loss that penalized the feature similarities of different samples in a mini-batch. Ulyanov \etal~\cite{ulyanov2017improved} minimized the Kullback-Leibler divergence between the generated distribution and a quasi-uniform distribution on the Julesz ensemble~\cite{julesz1981textons,zhu2000exploring}. Although their methods could generate diverse texture samples or stylized images to a certain extent, they still suffer from three main limitations. (1) Restricted generalization. Once trained, their feed-forward network is tied to a specific style, which cannot be generalized to other styles. (2) Limited diversity. Since their diversity is learned by penalizing the variations in mini-batches of a finite dataset and the weight of diversity loss should be set to a small value, the degree of diversity is limited. (3) Poor scalability. Extending their approaches to other methods requires the intractable modifications to training strategies and network structures, which might be useful for some learning-based methods like~\cite{huang2017arbitrary}, but not suitable for recent learning-free methods~\cite{li2017universal,sheng2018avatar,li2018closed} as these methods transfer arbitrary styles in a style-agnostic manner.

Facing the aforementioned challenges, we rethink the problem of diversity and an important insight we will use is that a Gram matrix~\cite{gatys2016image}, which is widely used as the style representation of an image, can correspond to an infinite number of different feature maps, and the images reconstructed from these feature maps are the diverse results we are looking for. Obviously, the problem of diversity has now been transformed into the problem of how to obtain the different feature maps with the same Gram matrix. Inspired by the work of Li \etal~\cite{li2017universal} which decomposes the Gram matrices and separates the matching of them by whitening and coloring transforms (WCTs), we propose a simple yet effective method, \ie, deep feature perturbation (DFP), to achieve diversified arbitrary style transfer. Our diversity is obtained by using an orthogonal noise matrix to perturb the image feature maps extracted from a DCNN while keeping the original style information unchanged. That is to say, although the perturbed feature maps are different from each other, they all have the same Gram matrix. For ease of understanding, we regard Gram matrix as the style representation, and define that different feature maps with the same Gram matrix share the same style-specific feature space.

In this work, our DFP is based on the framework of WCT~\cite{li2017universal}, so it can be easily incorporated into many WCT-based methods~\cite{li2017universal,sheng2018avatar,li2018closed} and empower them to generate diverse results without any extra learning process. Note that this learning-free process is fundamentally different from the aforementioned diversified methods that require learning with pre-defined styles. Therefore, our method is able to achieve diversified arbitrary style transfer.

The main contributions of this work are threefold:

•	We propose to use deep feature perturbation, \ie, perturbing the deep image feature maps by an orthogonal noise matrix while keeping the original style information unchanged, to achieve diversified arbitrary style transfer.

•	Our method can be easily incorporated into existing WCT-based methods~\cite{li2017universal,sheng2018avatar,li2018closed} which are used for different style transfer tasks, \eg, artistic style transfer, semantic-level style transfer and photo-realistic style transfer.

•	Theoretical analysis proves the capability of the proposed method in generating diversity, and the experimental results demonstrate that our method can greatly increase the diversity while maintaining the quality of stylization. 

\section{Related Work}

{\bf Gram-based Methods.} Gatys \etal~\cite{gatys2015neural,gatys2015texture,gatys2016image} first proposed an algorithm for arbitrary style transfer and texture synthesis based on matching the correlations (\ie, Gram matrix) between deep feature maps extracted from a pre-trained DCNN within an iterative optimization framework, but one major drawback is the inefficiency. To address this, Johnson \etal~\cite{johnson2016perceptual} and Ulyanov \etal~\cite{ulyanov2016texture,ulyanov2017improved} directly trained feed-forward generative networks for fast style transfer, but these methods need to retrain the network every time for a new style, which is inflexible. For this limitation, some methods~\cite{dumoulin2017learned,zhang2018multi,chen2017stylebank,li2017diversified,shen2018neural} were proposed to incorporate multiple styles into one single network, but they are still limited in a fixed number of pre-defined styles. More recently, Huang and Belongie~\cite{huang2017arbitrary} further allowed arbitrary style transfer in one single feed-forward network.

{\bf WCT-based Methods.} Recently, Li \etal~\cite{li2017universal} have proposed to exploit a series of feature transforms to achieve fast arbitrary style transfer in a style learning-free manner. They reformulated the task of style transfer as an image reconstruction process, with the feature maps of the content image being {\em whitened} at intermediate layers with regard to their style statistics (\ie, Gram matrix), and then {\em colored} to exhibit the same statistical characteristics of the style image. This method is essentially a Gram-based method, but it splits the Gram matrices by matrix decomposition, and separates the matching of them by whitening and coloring transforms (WCTs), thus providing an opportunity for our deep feature perturbation. Furthermore, Sheng \etal~\cite{sheng2018avatar} combined it with style swap~\cite{chen2016fast} for higher quality semantic-level style transfer. Li \etal~\cite{li2018closed} and Yoo \etal~\cite{yoo2019photorealistic} developed this to fast photo-realistic style transfer. More recently, Li \etal~\cite{li2018learning} derived the form of transformation matrix theoretically and directly learned it with a feed-forward network. Lu \etal~\cite{lu2019closed} derived a closed-form solution by treating it as the optimal transport problem. In our work, taking the most representative ones~\cite{li2017universal,sheng2018avatar,li2018closed} as examples, the proposed method can be easily integrated into the learning-free WCT process and empower these methods to generate diverse results, which will be shown in Section~\ref{exp}.

{\bf Diversified Methods.} Our method is closely related to~\cite{li2017diversified} and~\cite{ulyanov2017improved}. Li \etal~\cite{li2017diversified} introduced a diversity loss to allow the feed-forward networks to generate diverse outputs. It explicitly measures the variations in visual appearances between the generated results, and penalizes them in a mini-batch. Ulyanov \etal~\cite{ulyanov2017improved} proposed a new formulation that allowed to train generative networks which sampled the Julesz ensemble~\cite{julesz1981textons,zhu2000exploring}. Specifically, the diversity term of its learning objective is similar to that of  Li \etal~\cite{li2017diversified}, which quantifies the lack of diversity in the batch by mutually comparing the generated images. Although these methods could generate diverse outputs to a certain extent, they still suffer from the restricted generalization, limited diversity and poor scalability, as we have introduced in Section~\ref{introduction}.

The proposed method is based on WCT~\cite{li2017universal}, and can be easily integrated into WCT-based methods to empower them to generate diverse results. Unlike the previous diversified methods~\cite{li2017diversified,ulyanov2017improved} that need to train an independent network for every style, our diversity is learning-free and suitable for arbitrary styles. Moreover, without extra constraints, our method can generate an infinite number of solutions with satisfactory quality as well as distinct diversity.
\begin{figure*}[htbp]
	\centering
	\includegraphics[width=1\linewidth]{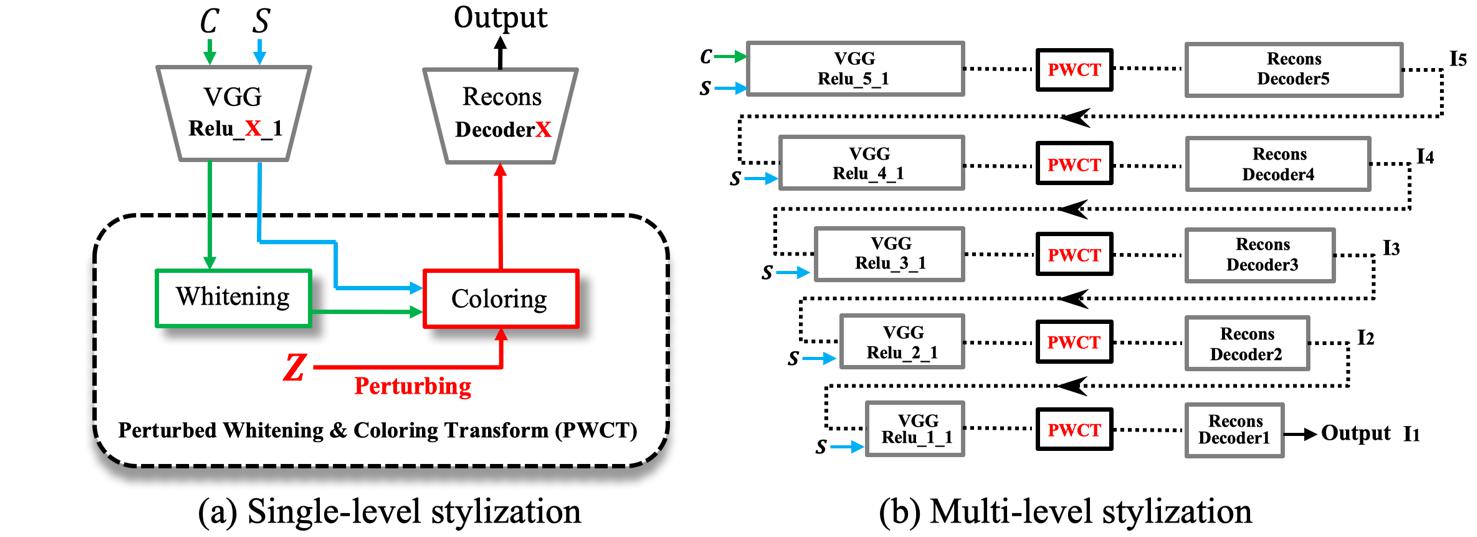}
	\caption{Our diversified arbitrary style transfer pipeline. (a) We add an orthogonal noise matrix ${\bf Z}$ to perturb the whitening and coloring transform (WCT). Like~\cite{li2017universal}, the VGG and DecoderX are first trained for image reconstruction and then fixed for style transfer. $C$ and $S$ denote the content image and style image, respectively. (b) Our perturbed whitening and coloring transform (PWCT) can be applied in every level of the multi-level stylization framework of~\cite{li2017universal}.
	}
	\label{fig:fig1}
\end{figure*}

\section{Style-Specific Feature Space }
\label{SSIS}
Defining the style of an image is a quite tricky problem, and so far no unified conclusion has been reached. Informally, a style can be regarded as a family of visual attributes, such as color, brush strokes and line drawing, etc. Recently, Gatys \etal~\cite{gatys2015neural,gatys2015texture,gatys2016image} have proposed a new style representation (Gram matrix) for artistic images. In their works, the style of an image is represented by the correlations between deep feature maps extracted from a pre-trained DCNN. Given an image $\vec{x}$ as input, the vectorized feature map extracted from a certain layer (we only take one layer as an example) of the VGG model~\cite{simonyan2014very} is denoted as $F\in \mathbb{R}^{C\times HW}$, where $H$, $W$ are the height and width of the original feature map, $C$ is the number of channels. The style of the image $\vec{x}$ can be represented as follows:
\begin{equation}
G_{ij}=\sum_{k}F_{ik}F_{jk}=FF^T\in \mathbb{R}^{C\times C},
\label{gram}
\end{equation}
where $F_{ik}$ and $F_{jk}$ are the activations of the $i^{th}$ and $j^{th}$ filter at position $k$, $F^T$ is the transpose matrix of $F$.

It is obvious that, for a definite Gram matrix $\mathcal{G}$, there could be a large number of feature maps corresponding to it. Let $\mathcal{F}_l$ denote the vectorized feature map of an image in layer $l$. $\mathcal{F}_l$ is perceived as the style $\mathcal{G}$ in layer $l$ if its Gram matrix matches $\mathcal{G}$. Formally, given the loss function:
\begin{equation}
\mathcal{L}_\mathcal{G}(\mathcal{F}_l)=||\mathcal{F}_l\mathcal{F}_l^T-\mathcal{G}||,
\end{equation}
we define the feature maps that satisfy the following constraint belong to the same style-specific feature space of $\mathcal{G}$.
\begin{equation}
\mathcal{S}_\mathcal{G}=\{\mathcal{F}_l\in \mathbb{F}: \mathcal{L}_\mathcal{G}(\mathcal{F}_l)=0\},
\label{sen}
\end{equation}
where $\mathbb{F}$ is a set of feature maps. Features belonging to the same $\mathcal{S}$ are perceptually equivalent in style characteristics.

In particular, sometimes we do not need their Gram matrices to be exactly equal, and then we can get the relaxed constraint,
\begin{equation}
\mathcal{S}_\mathcal{G}^{\epsilon}=\{\mathcal{F}_l\in \mathbb{F}: \mathcal{L}_\mathcal{G}(\mathcal{F}_l)\leq \epsilon\},
\label{rsen}
\end{equation}
in which the feature maps are approximately equivalent in style characteristics.

In this work, our deep feature perturbation can easily achieve the first constraint (Eq.~(\ref{sen})), while the methods \cite{li2017diversified,ulyanov2017improved} only satisfy the second constraint (Eq.~(\ref{rsen})). That is to say, the Gram matrices of the diverse perturbed feature maps obtained by our method can be completely equal.

\section{Deep Feature Perturbation}
\label{DFP}
Our deep feature perturbation (DFP) is based on the work of Li \etal~\cite{li2017universal} and incorporated into its whitening and coloring transform (WCT) process to help generate diverse stylized results. The pipeline of our method is shown in Fig.~\ref{fig:fig1}, where the diversified style transfer is mainly achieved by the perturbed whitening and coloring transform (PWCT), which consists of two steps, \ie, whitening transform and perturbed coloring transform.

\begin{table*}
	\caption{Quantitative comparisons between single-level perturbation and multi-level perturbation in terms of run-time, tested on images of size $512\times 512$ and a 6GB Nvidia 980Ti GPU.}
	\centering
	\hspace{2cm}
	\setlength{\tabcolsep}{0.25cm}
	\begin{tabular}{ccccccccccc}
		\toprule
		Fig.~\ref{fig:fig2}&Li \etal~\cite{li2017universal}&I5&I4&I3&I2&I1&I5+I4&I5+I1&I3+I2+I1&I5+I4+I3+I2+I1\\
		\midrule 
		Time/sec&3.01&{\bf 3.53}&3.51&3.04&3.03&3.02&4.14&3.54&3.05&4.15\\
		\toprule
		Fig.~\ref{fig:fig3}&Li \etal~\cite{li2018closed}&-&I4&I3&I2&I1&I4+I3&I4+I1&I2+I1&I4+I3+I2+I1\\
		\midrule
		Time/sec&0.29&-&{\bf 0.32}&0.31&0.30&0.29&0.33&0.32&0.30&0.34\\
		\bottomrule
	\end{tabular}
	\label{tab1}
\end{table*}

{\bf Whitening Transform.} Given a pair of content image $I_c$ and style image $I_s$, we first extract their vectorized VGG feature maps $F_c=\Phi(I_c)\in \mathbb{R}^{C\times H_cW_c}$ and $F_s=\Phi(I_s)\in \mathbb{R}^{C\times H_sW_s}$ at a certain layer $\Phi$ (\eg, $Relu\_3\_1$), where $H_c$, $W_c$ ($H_s$, $W_s$) are the height and width of the content (style) feature, and $C$ is the number of channels. We first center $F_c$ by subtracting its mean vector $m_c$. Then the whitening transform (Eq.~(\ref{eq5})) is used to transform $F_c$ to $\hat{F_c}$, in which the feature maps are uncorrelated from each other (\ie, $\hat{F_c}\hat{F_c}^T=I$).

\renewcommand\arraystretch{0.5}
\begin{figure}[t]
	\centering
	\setlength{\tabcolsep}{0.01cm}
	\begin{tabular}{cp{0.05cm}ccccc}
		\includegraphics[width=0.161\linewidth]{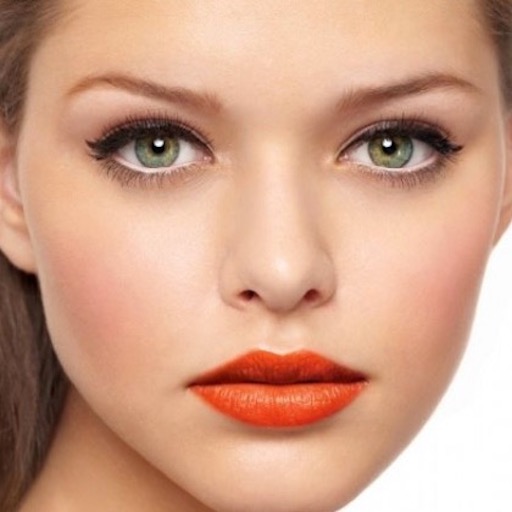}&&
		\includegraphics[width=0.161\linewidth]{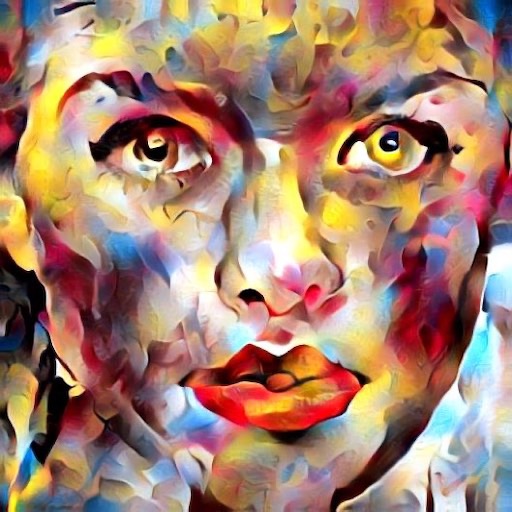}&
		\includegraphics[width=0.161\linewidth]{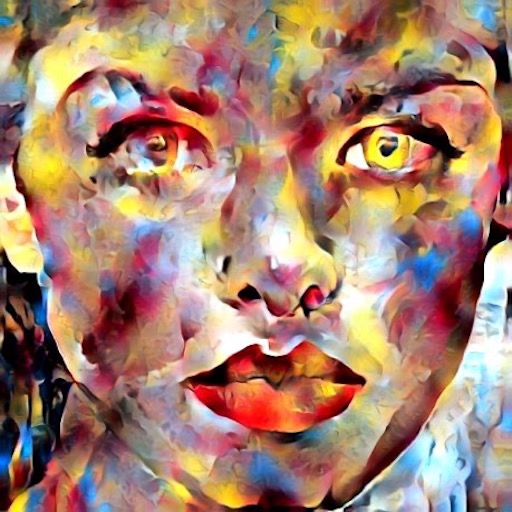}&
		\includegraphics[width=0.161\linewidth]{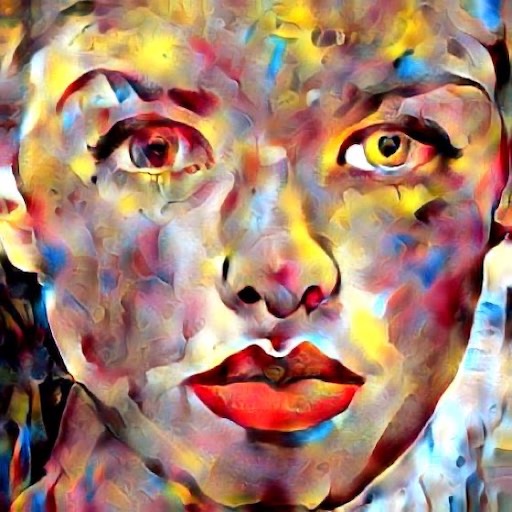}&
		\includegraphics[width=0.161\linewidth]{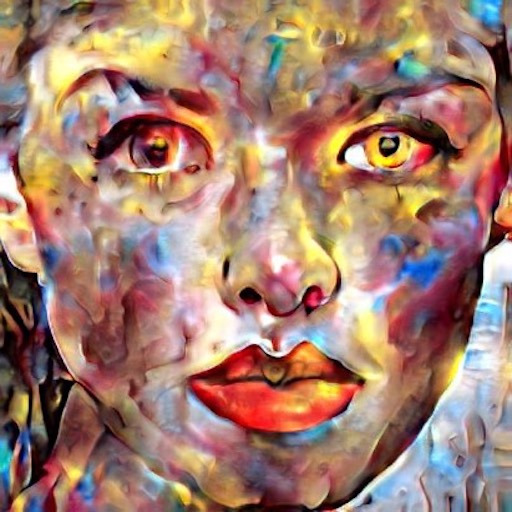}&
		\includegraphics[width=0.161\linewidth]{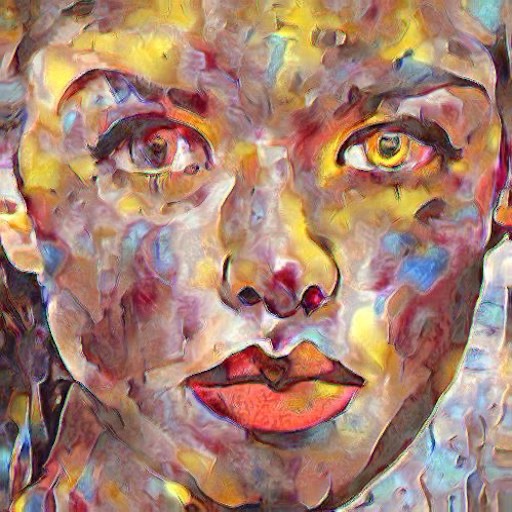}\\
		\scriptsize Content&&\scriptsize I5&\scriptsize I4&\scriptsize I3&\scriptsize I2&\scriptsize I1\\
		
		\includegraphics[width=0.161\linewidth]{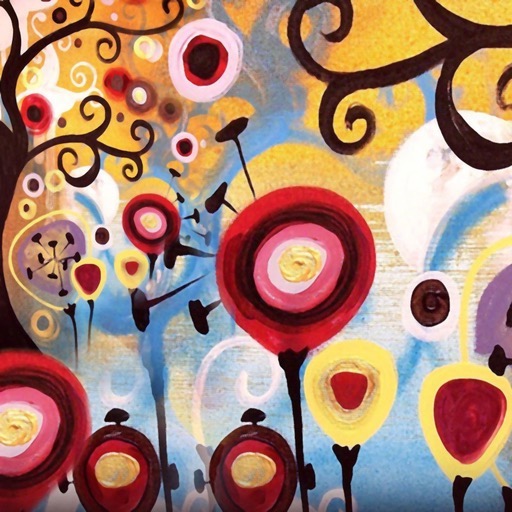}&&
		\includegraphics[width=0.161\linewidth]{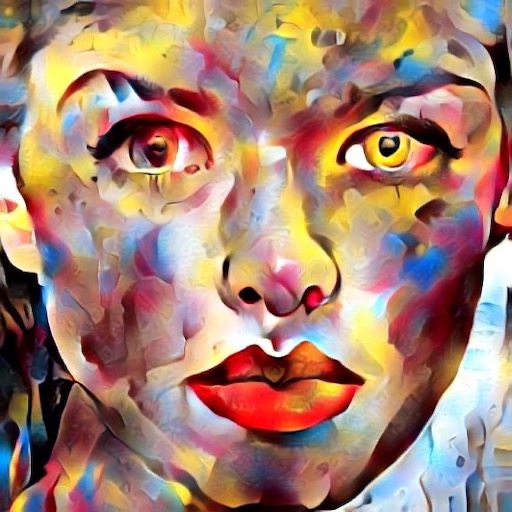}&
		\includegraphics[width=0.161\linewidth]{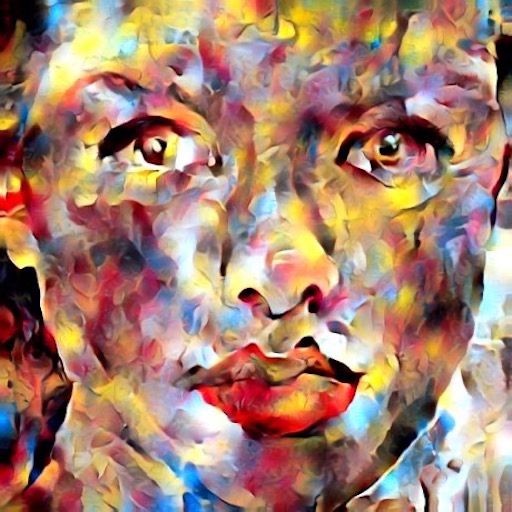}&
		\includegraphics[width=0.161\linewidth]{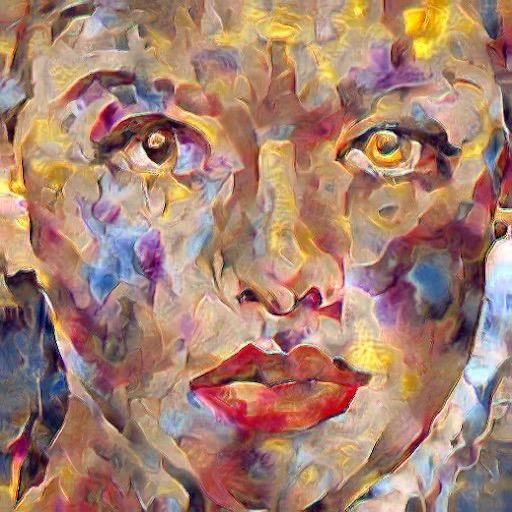}&
		\includegraphics[width=0.161\linewidth]{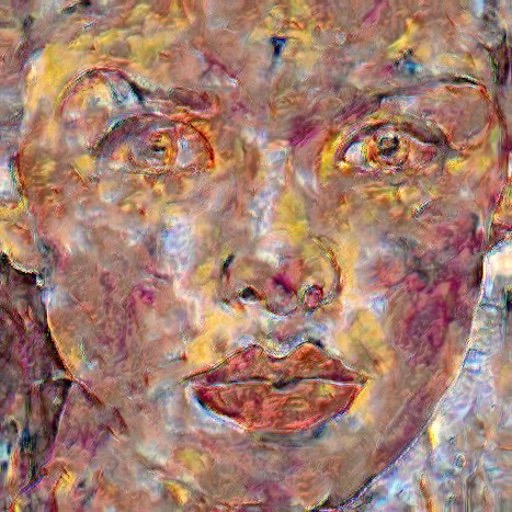}&
		\includegraphics[width=0.161\linewidth]{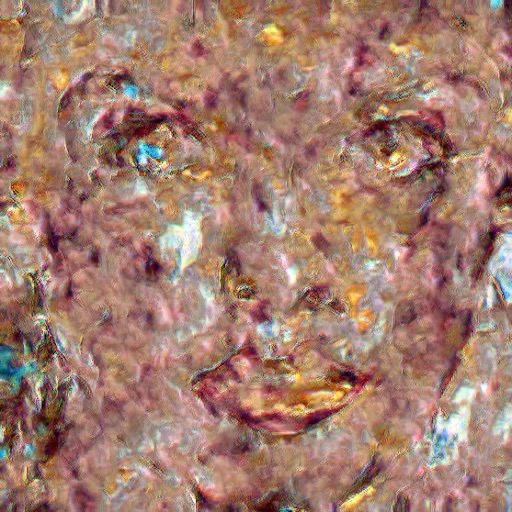}\\
		\scriptsize Style&&\scriptsize Li~\etal~\cite{li2017universal}&\scriptsize I5 + I4&\scriptsize I5 + I1&\scriptsize I3 + I2 + I1&\fontsize{5.8pt}{\baselineskip}\selectfont{I5+I4+I3+I2+I1}\\
	\end{tabular}
	\caption{Single-level perturbation vs. Multi-level perturbation. Our DFP is integrated into method~\cite{li2017universal}. The top row shows the results obtained by only perturbing a single-level stylization in Fig.~\ref{fig:fig1}(b). The bottom row shows the results obtained by perturbing stylizations in multiple levels.
    }
    \label{fig:fig2}
\end{figure}

\begin{equation}
\hat{F_c}=E_cD_c^{-\frac{1}{2}}E_c^TF_c,
\label{eq5}
\end{equation}
where $D_c$ and $E_c$ are obtained by the singular value decomposition (SVD) of the Gram matrix $F_cF_c^T\in\mathbb{R}^{C\times C}$ (Eq.~(\ref{gram})), \ie, $F_cF_c^T=E_cD_cE_c^T$. $D_c$ is the diagonal matrix of the eigenvalues, and $E_c$ is the corresponding orthogonal matrix of eigenvectors.

{\bf Perturbed Coloring Transform.} We first center $F_s$ by subtracting its mean vector $m_s$. The coloring transform used in~\cite{li2017universal} is essentially the inverse of the whitening step, \ie, using Eq.~(\ref{eq6}) to transform $\hat{F_c}$ so that we can obtain $\hat{F_{cs}}$ which satisfies the same Gram matrix of $F_s$ (\ie, $\hat{F_{cs}}\hat{F_{cs}}^T=F_sF_s^T$).
\begin{equation}
\hat{F_{cs}}=E_sD_s^{\frac{1}{2}}E_s^T\hat{F_c},
\label{eq6}
\end{equation}
where $D_s$ and $E_s$ are obtained by the SVD of the Gram matrix $F_sF_s^T\in\mathbb{R}^{C\times C}$, \ie, $F_sF_s^T=E_sD_sE_s^T$. $D_s$ is the diagonal matrix of the eigenvalues, and $E_s$ is the corresponding orthogonal matrix of eigenvectors.

The goal of coloring transform is to make the Gram matrix of $\hat{F_{cs}}$ the same as that of $F_s$. According to our analysis in Section~\ref{SSIS}, these two feature maps share the same style-specific feature space. In theory, $\hat{F_{cs}}$ should have a large number of possibilities, but Eq.~(\ref{eq6}) only produces one of them. In order to traverse these solutions as much as possible, we propose to use deep feature perturbation.

\renewcommand\arraystretch{0.5}
\begin{figure}[t]
	\centering
	\setlength{\tabcolsep}{0.01cm}
	\begin{tabular}{cp{0.05cm}cccc}
		\includegraphics[width=0.195\linewidth]{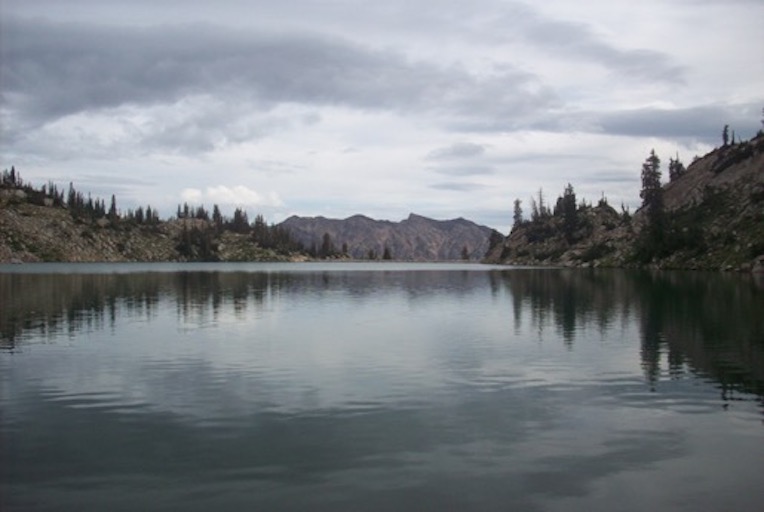}&&
		\includegraphics[width=0.195\linewidth]{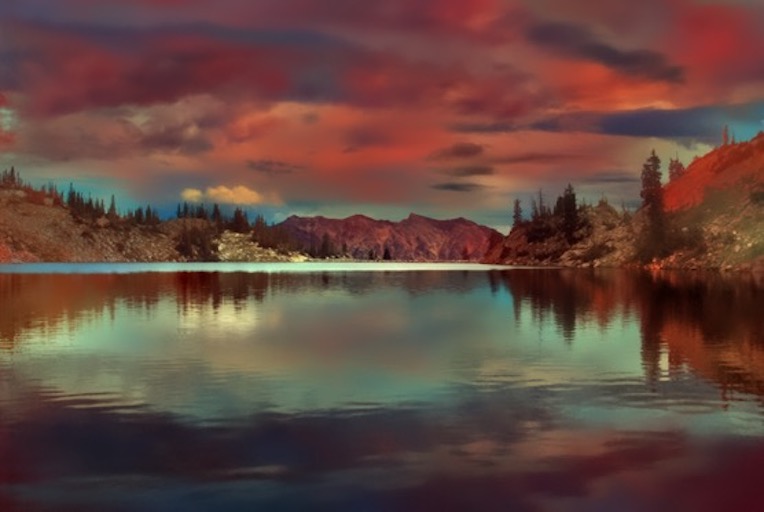}&
		\includegraphics[width=0.195\linewidth]{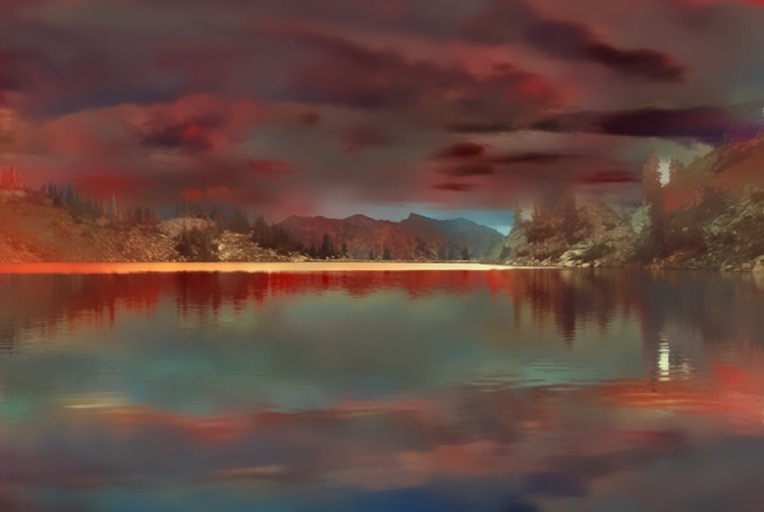}&
		\includegraphics[width=0.195\linewidth]{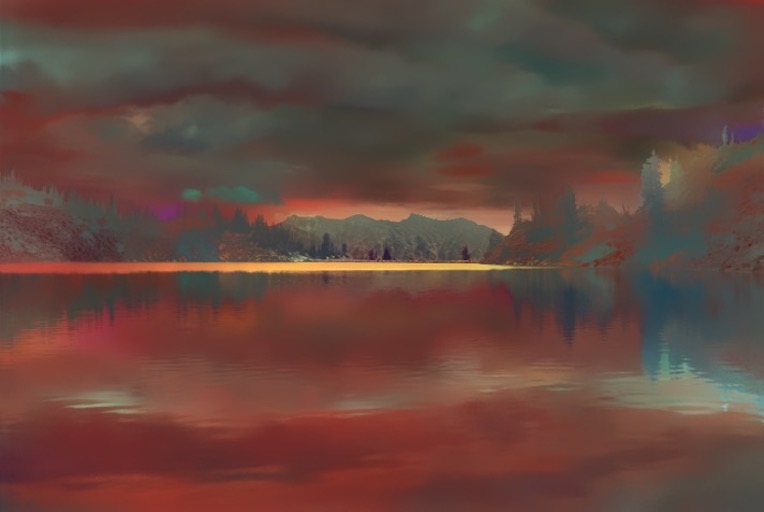}&
		\includegraphics[width=0.195\linewidth]{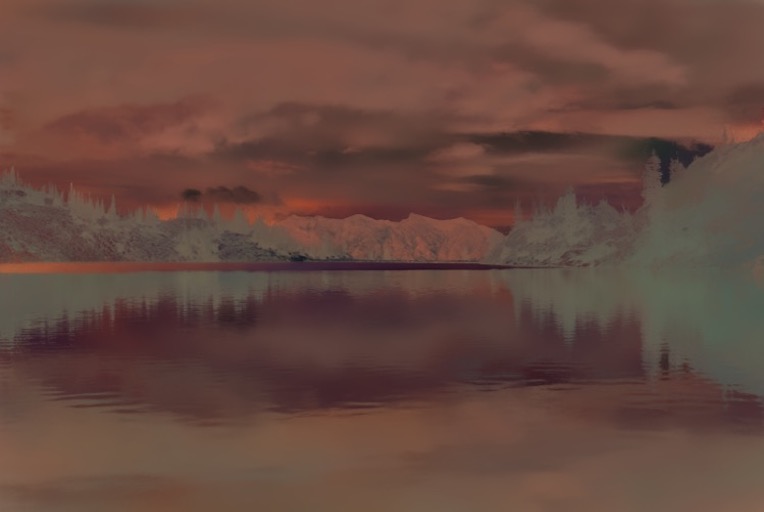}\\
		\scriptsize Content&&\scriptsize I4&\scriptsize I3&\scriptsize I2&\scriptsize I1\\
		
		\includegraphics[width=0.195\linewidth]{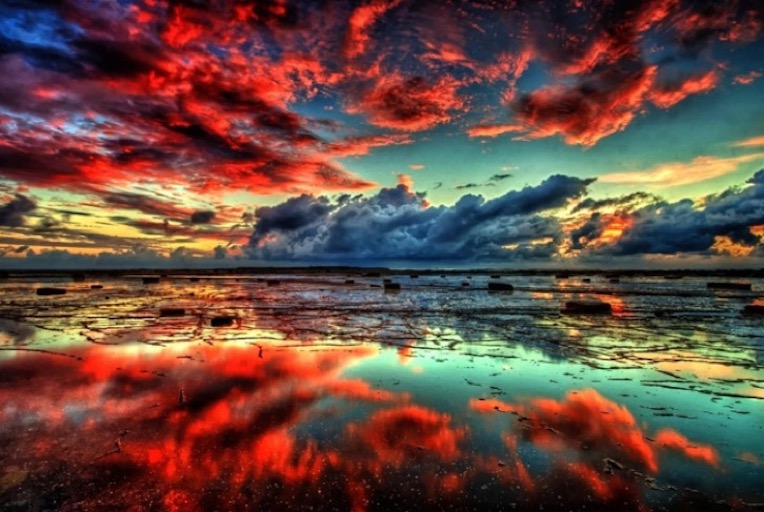}&&
		\includegraphics[width=0.195\linewidth]{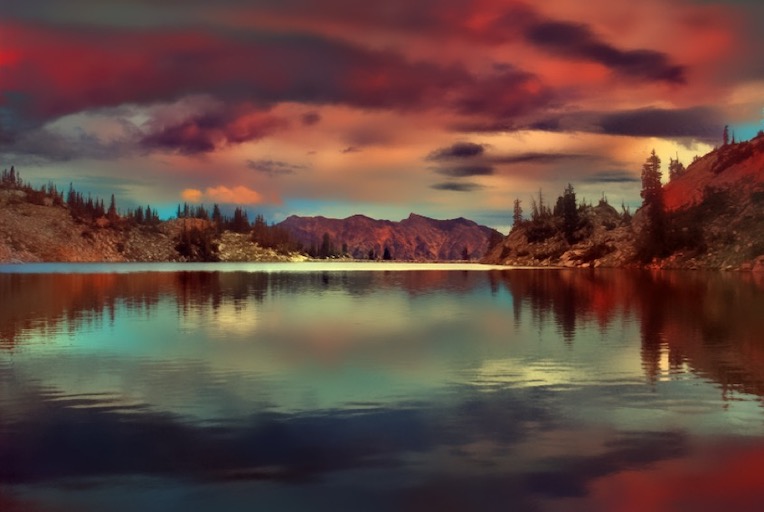}&
		\includegraphics[width=0.195\linewidth]{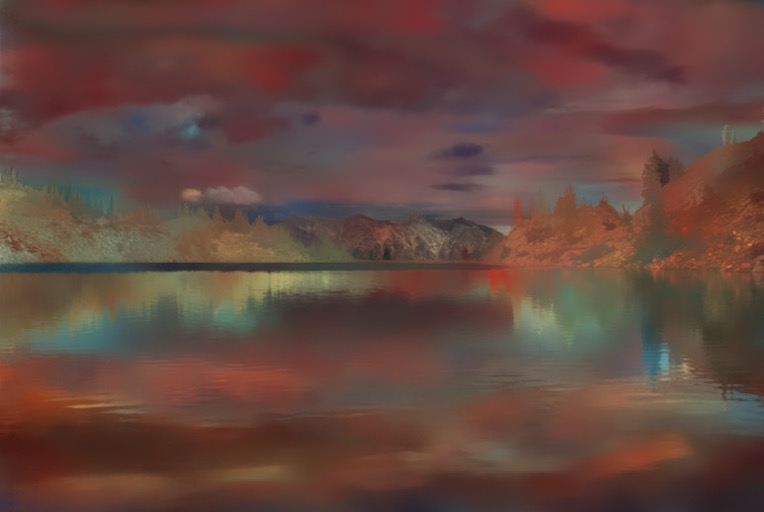}&
		\includegraphics[width=0.195\linewidth]{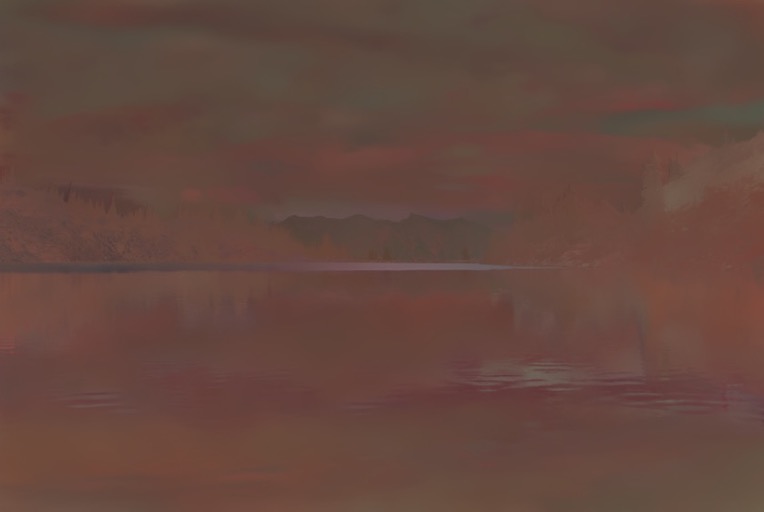}&
		\includegraphics[width=0.195\linewidth]{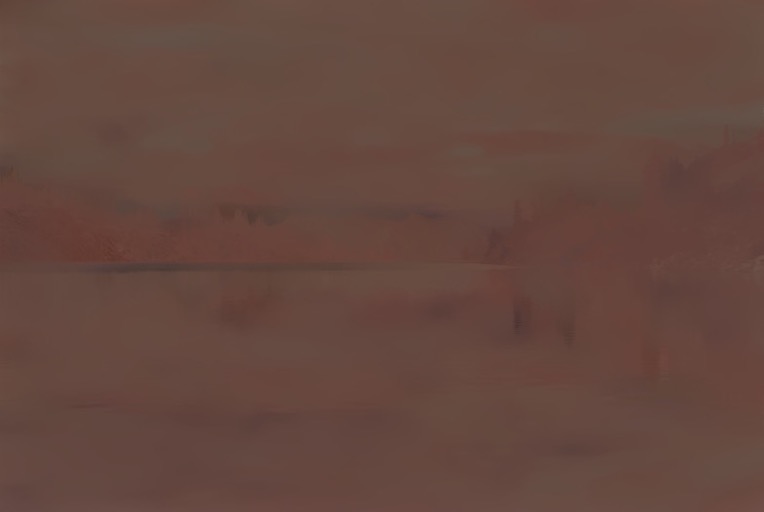}\\
		\scriptsize Style&&\scriptsize Li~\etal~\cite{li2018closed}&\scriptsize I4 + I3&\scriptsize I4 + I1&\scriptsize I4 + I3 + I2 + I1\\
	\end{tabular}
	\caption{Another comparison of Single-level and Multi-level perturbation. Our DFP is integrated into method~\cite{li2018closed}. This method only uses four-level stylizations. The top row shows the results obtained by only perturbing a single-level stylization. The bottom row shows the results obtained by perturbing stylizations in multiple levels.
	}
	\label{fig:fig3}
\end{figure}

The key idea of our deep feature perturbation is incorporating an orthogonal noise matrix into Eq.~(\ref{eq6}) to perturb the feature $\hat{F_{cs}}$ while preserving its Gram matrix. Obviously, there are three places to insert the noise matrix, \ie, between $D_s^{\frac{1}{2}}$ and $E_s^T$, between $E_s^T$ and $\hat{F_c}$, and on the right side of $\hat{F_c}$ (since $E_s^TE_s=I$ and $\hat{F_c}\hat{F_c}^T=I$). We eventually insert the orthogonal noise matrix between $D_s^{\frac{1}{2}}$ and $E_s^T$ as this may consume the least computation and run-time (we will discuss this in Section~\ref{abstudy}).

We first obtain a random noise matrix $N$ (\eg, sampled from the standard normal distribution, we will discuss it in Section~\ref{abstudy}) according to the shape of $D_s^{\frac{1}{2}}$ and $E_s^T$. Assume that the shape of $D_s^{\frac{1}{2}}$ is $(C-k)\times(C-k)$, where $k$ is the number of small singular values (\eg, less than $10^{-5}$, Li \etal~\cite{li2017universal} suggest removing these small singular values to obtain higher quality results), and the shape of $E_s^T$ is $(C-k)\times C$, then the shape of $N$ is $(C-k)\times(C-k)$. To obtain orthogonal noise matrix, we apply the SVD to decompose $N$, \ie, $N=E_nD_nV_n^T$, and directly use the orthogonal matrix ${\bf Z}=E_n\in \mathbb{R}^{(C-k)\times (C-k)}$. Finally, we insert ${\bf Z}$ between $D_s^{\frac{1}{2}}$ and $E_s^T$ of Eq.~(\ref{eq6}). Our new perturbed coloring transform is formulated as follows:
\begin{equation}
\hat{F_{csn}}=E_sD_s^{\frac{1}{2}}{\bf Z}E_s^T\hat{F_c},
\label{eq7}
\end{equation}
since ${\bf ZZ}^T=I$, we can deduce as follows:
$$\hat{F_{csn}}\hat{F_{csn}}^T=(E_sD_s^{\frac{1}{2}}{\bf Z}E_s^T\hat{F_c})(\hat{F_c}^TE_s{\bf Z^T}D_s^{\frac{1}{2}}E_s^T)$$

$=E_sD_s^{\frac{1}{2}}({\bf Z}E_s^T\hat{F_c}\hat{F_c}^TE_s{\bf Z^T})D_s^{\frac{1}{2}}E_s^T$
\vspace{8pt}

$=E_sD_sE_s^T=F_sF_s^T$
\vspace{8pt}

\renewcommand\arraystretch{0.5}
\begin{figure}[t]
	\centering
	\setlength{\tabcolsep}{0.03cm}
	\begin{tabular}{cccccc}
		\includegraphics[width=0.16\linewidth]{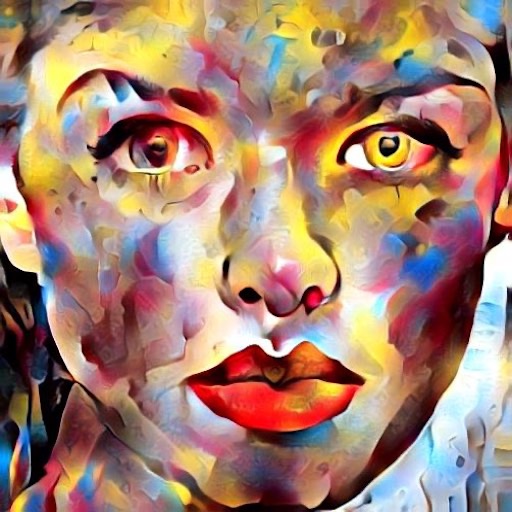}&
		\includegraphics[width=0.16\linewidth]{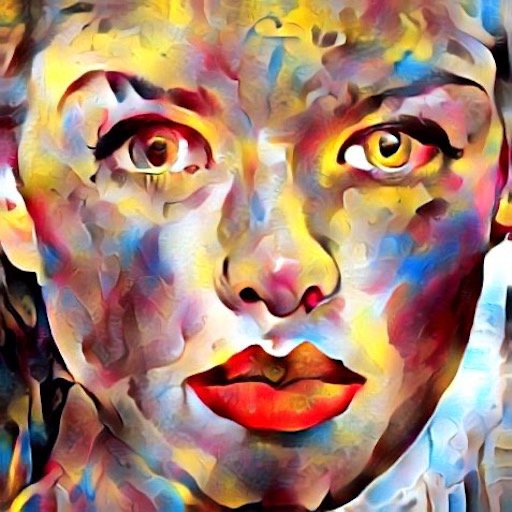}&
		\includegraphics[width=0.16\linewidth]{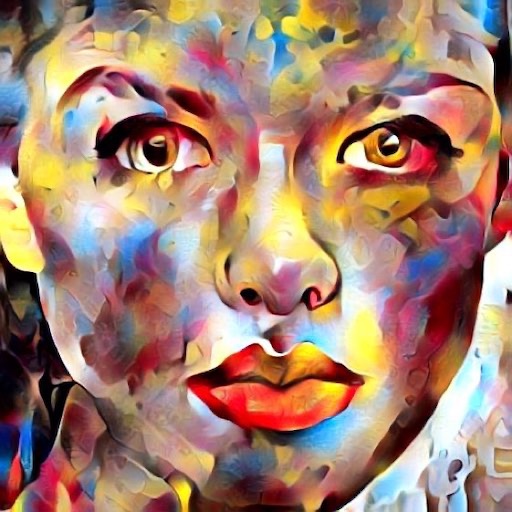}&
		\includegraphics[width=0.16\linewidth]{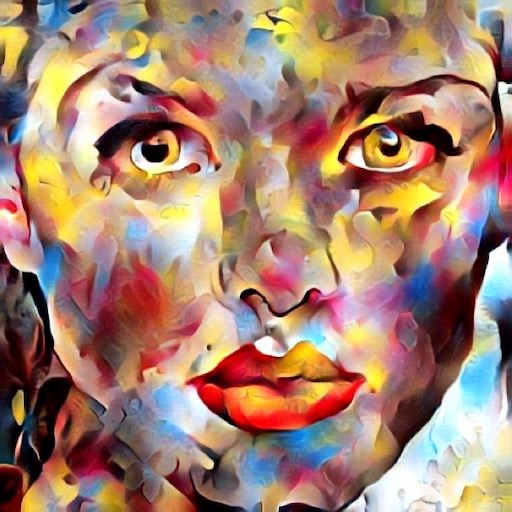}&
		\includegraphics[width=0.16\linewidth]{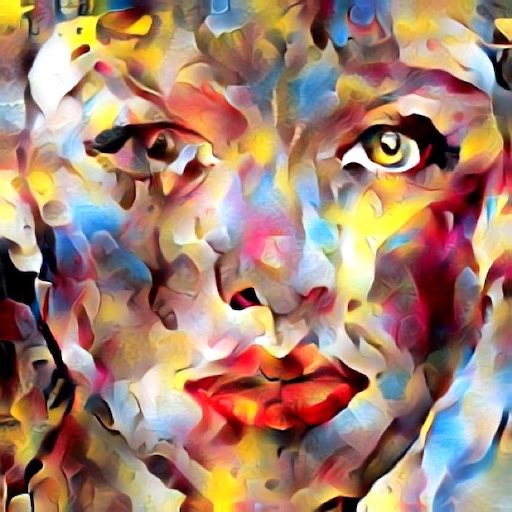}&
		\includegraphics[width=0.16\linewidth]{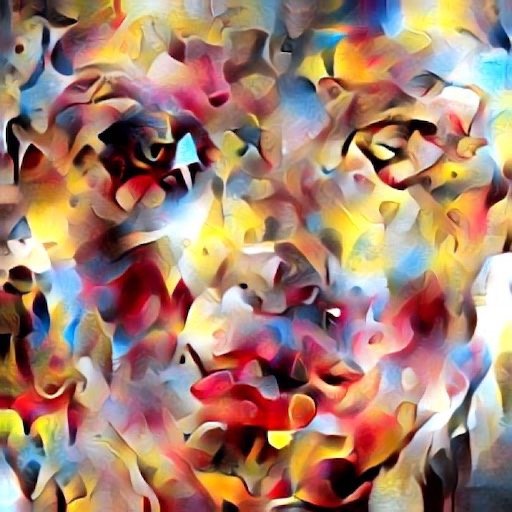}\\
		
		\includegraphics[width=0.16\linewidth]{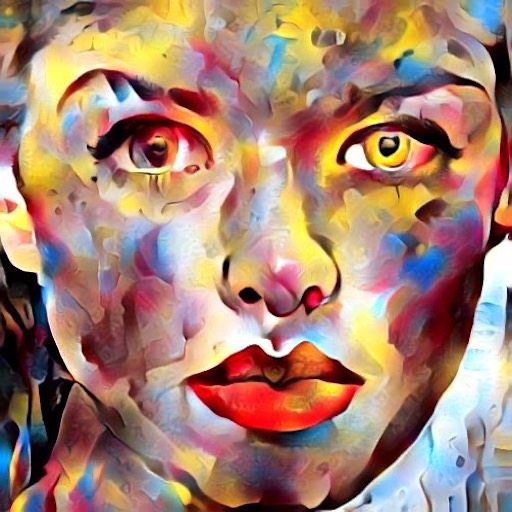}&
		\includegraphics[width=0.16\linewidth]{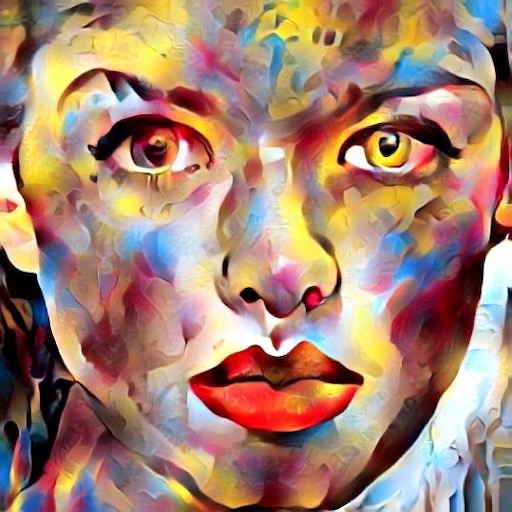}&
		\includegraphics[width=0.16\linewidth]{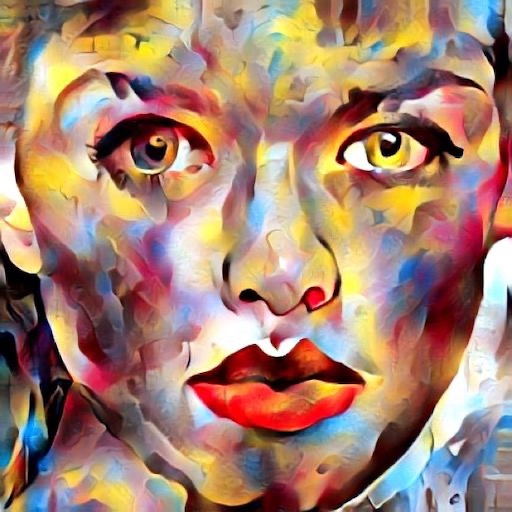}&
		\includegraphics[width=0.16\linewidth]{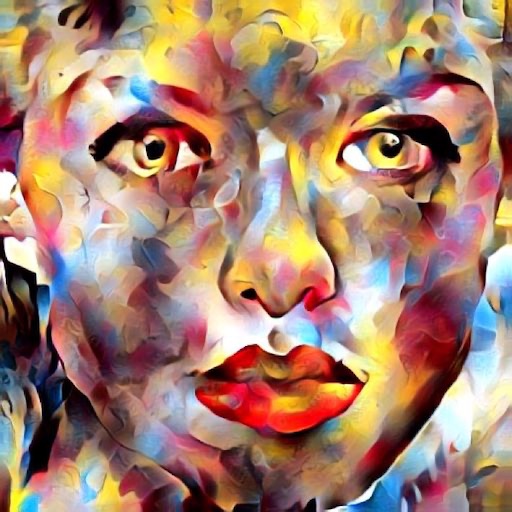}&
		\includegraphics[width=0.16\linewidth]{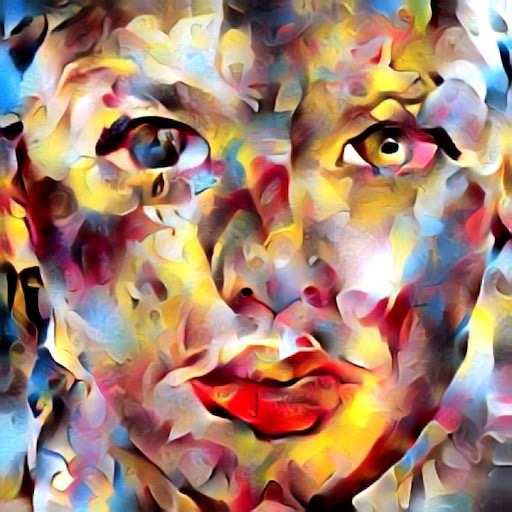}&
		\includegraphics[width=0.16\linewidth]{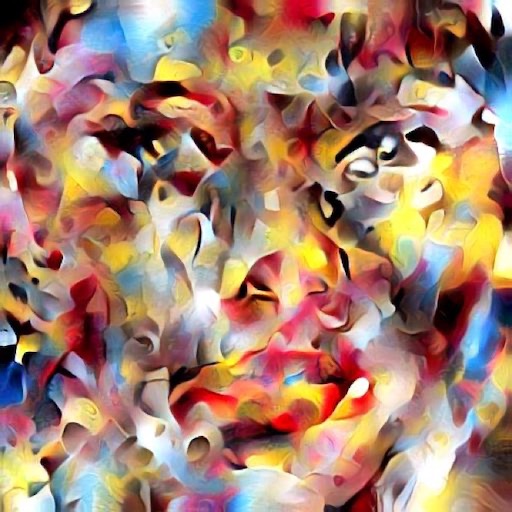}\\
		
		\scriptsize $\lambda=0$&\scriptsize $\lambda=0.2$&\scriptsize $\lambda=0.4$&\scriptsize $\lambda=0.6$&\scriptsize $\lambda=0.8$&\scriptsize $\lambda=1.0$\\
	\end{tabular}
	\caption{Trade-off between diversity and quality by varying diversity hyperparameter $\lambda$ in method~\cite{li2017universal} (+ our DFP).
	}
	\label{fig:fig4}
\end{figure}
In our later experiments, we find that only using our perturbed coloring transform may reduce the quality of stylization. This may be because $\hat{F_{cs}}$ (Eq. (\ref{eq6})) contains not only style information (Gram matrix) from $E_sD_s^{\frac{1}{2}}E_s^T$, but also some {\em content information from} $\hat{F_c}$ (Eq. (\ref{eq5})). Although our feature perturbation (Eq. (\ref{eq7})) can keep the style information unchanged, the content information may be affected by the noise matrix, which is manifested as a decline in quality. Fortunately, in WCT-based methods~\cite{li2017universal,sheng2018avatar,li2018closed}, the content information in $\hat{F_c}$ is {\em not the determinant} of the content in the final result, as in these methods $\hat{F_{cs}}$ is mainly served as the style feature, and blended with the content feature $F_c$ to balance the style and content (similar to our Eq. (\ref{eq9})). In order to increase the diversity while maintaining the original quality, we introduce a diversity hyperparameter $\lambda$ to provide user controls on the trade-off between them.
\begin{equation}
\hat{F_{csn}}'= \lambda \hat{F_{csn}}+(1-\lambda)\hat{F_{cs}}.
\label{eq8}
\end{equation}

Then, we re-center the $\hat{F_{csn}}'$ with the mean vector $m_s$ of the style, \ie, $\hat{F_{csn}}'=\hat{F_{csn}}'+m_s$. At last, we blend $\hat{F_{csn}}'$ with the content feature $F_c$ before feeding it to the decoder.
\begin{equation}
\hat{F_{csn}}'= \alpha \hat{F_{csn}}'+(1-\alpha)F_c,
\label{eq9}
\end{equation}
where the hyperparameter $\alpha$ serves as the weight for users to control the stylization strength, like~\cite{li2017universal}.

{\bf Multi-level Stylization.} We follow the multi-level coarse-to-fine stylization used in~\cite{li2017universal}, but replace their WCTs with our PWCTs, as shown in Fig.~\ref{fig:fig1} (b). In fact, we do not need to add noise to every level. We will discuss this in Section~\ref{abstudy}.

{\bf Discussions.} As a matter of fact, optimizing the diversity loss of~\cite{li2017diversified,ulyanov2017improved} can be viewed as a sub-optimal approximation of our method, as analyzed in Section~\ref{SSIS}. But since the diversity loss is only optimized on mini-batches of a finite dataset and the weight should be set to a small value (otherwise it will seriously reduce the quality), the degree of diversity is limited. By contrast, the different orthogonal noise matrices can be innumerable and diverse, so there could be endless possibilities with distinct diversity for the results of our approach. Moreover, our method is learning-free and can be effective for arbitrary styles, while the diversity loss of~\cite{li2017diversified,ulyanov2017improved} needs to be optimized every time for every style.

\renewcommand\arraystretch{0.5}
\begin{figure}[t]
	\centering
	\setlength{\tabcolsep}{0.03cm}
	\begin{tabular}{cccccc}
		\includegraphics[width=0.16\linewidth]{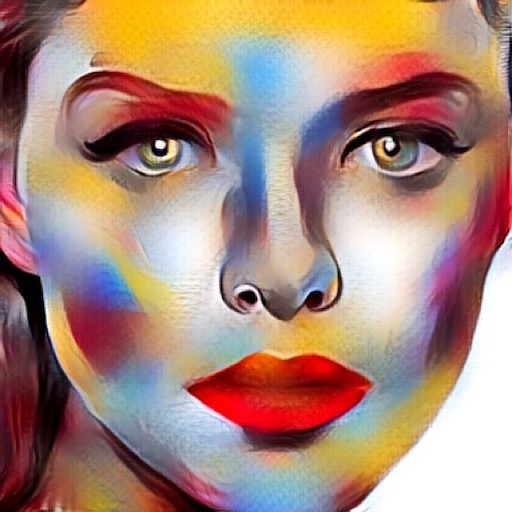}&
		\includegraphics[width=0.16\linewidth]{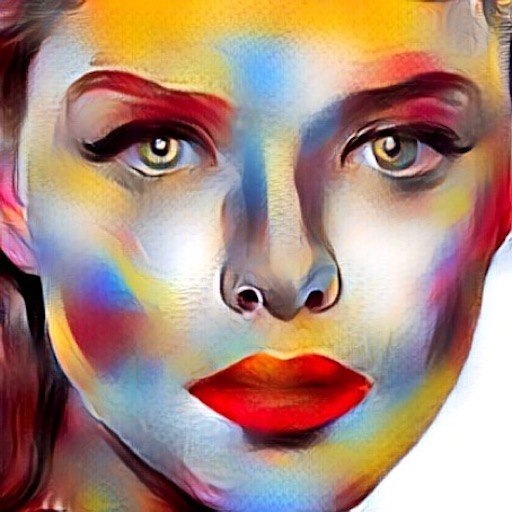}&
		\includegraphics[width=0.16\linewidth]{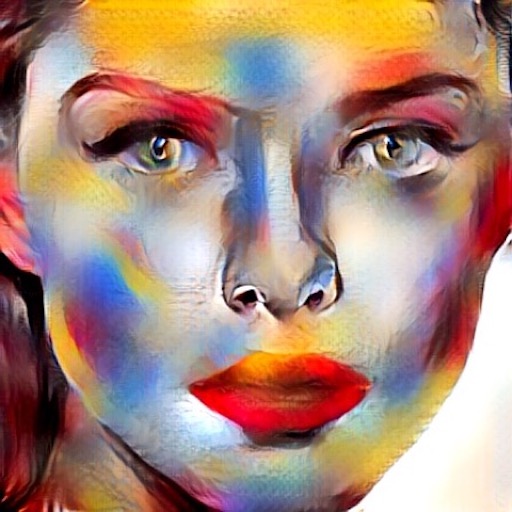}&
		\includegraphics[width=0.16\linewidth]{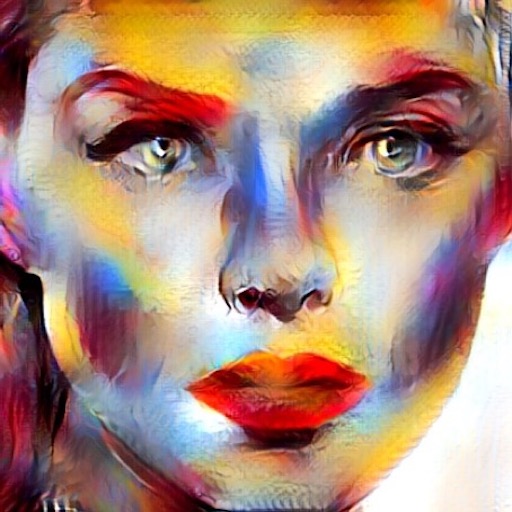}&
		\includegraphics[width=0.16\linewidth]{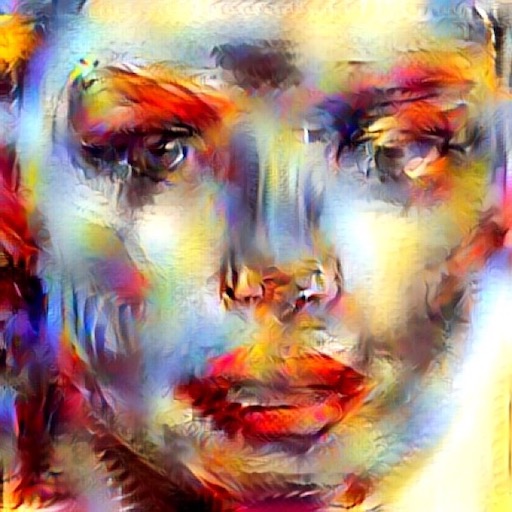}&
		\includegraphics[width=0.16\linewidth]{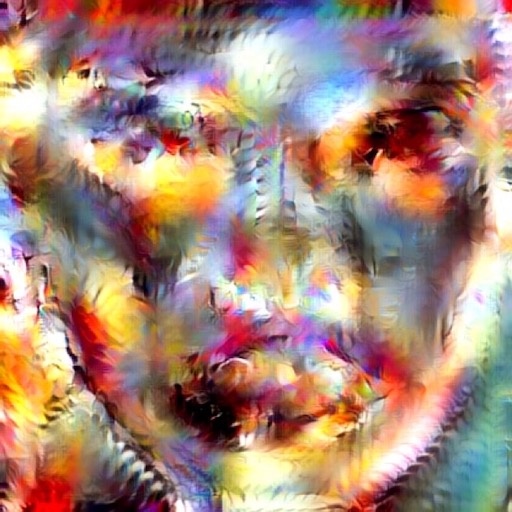}\\
		
		\includegraphics[width=0.16\linewidth]{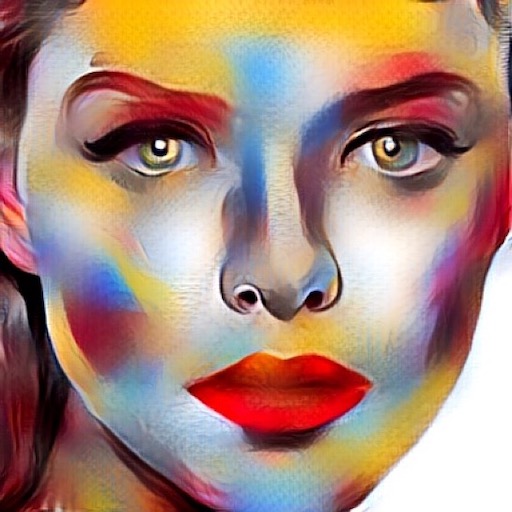}&
		\includegraphics[width=0.16\linewidth]{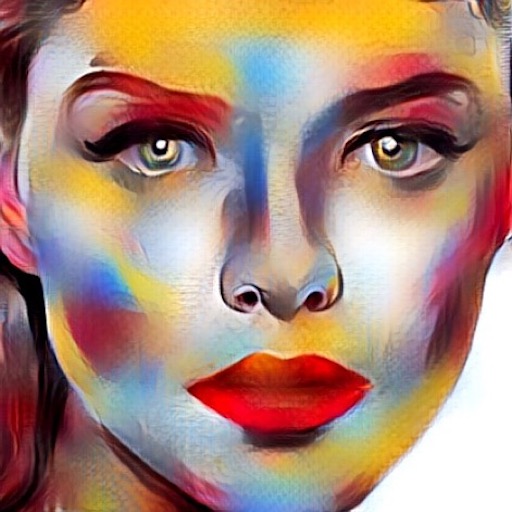}&
		\includegraphics[width=0.16\linewidth]{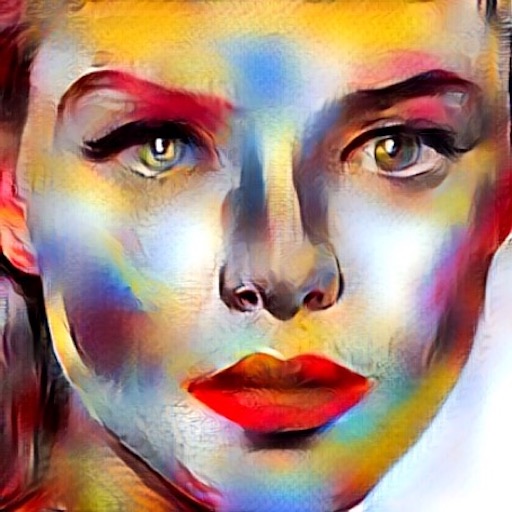}&
		\includegraphics[width=0.16\linewidth]{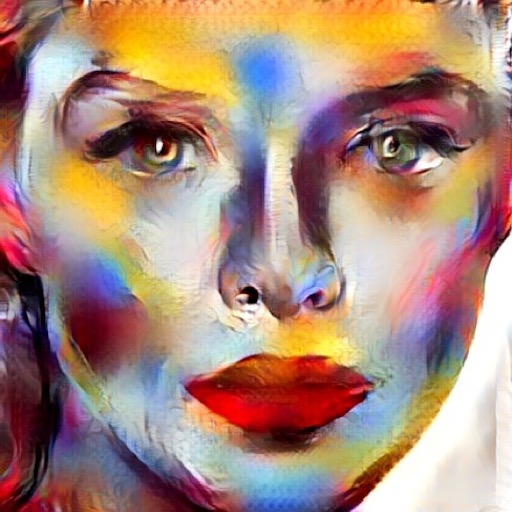}&
		\includegraphics[width=0.16\linewidth]{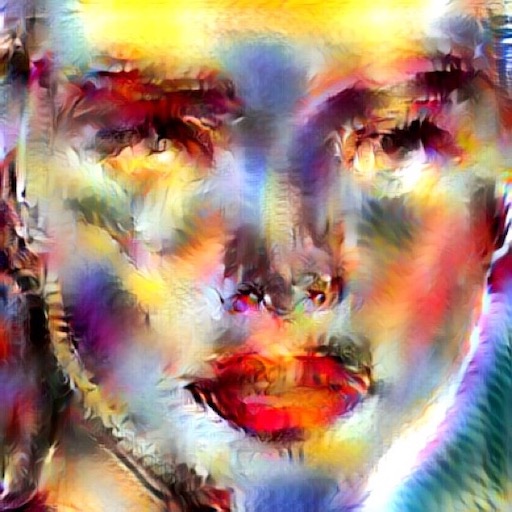}&
		\includegraphics[width=0.16\linewidth]{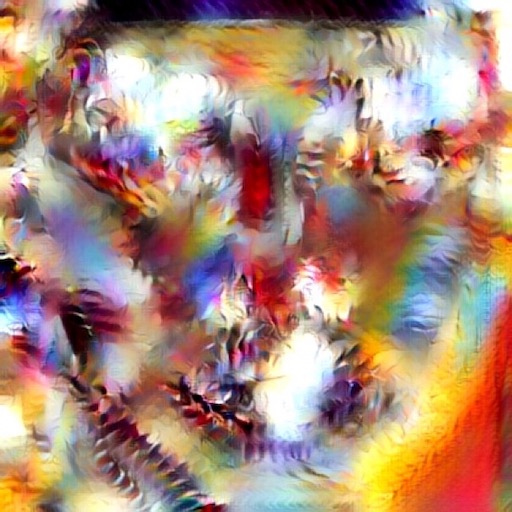}\\
		
		\scriptsize $\lambda=0$&\scriptsize $\lambda=0.3$&\scriptsize $\lambda=0.5$&\scriptsize $\lambda=0.6$&\scriptsize $\lambda=0.8$&\scriptsize $\lambda=1.0$\\
	\end{tabular}
	\caption{Trade-off between diversity and quality by varying diversity hyperparameter $\lambda$ in method~\cite{sheng2018avatar} (+ our DFP).
	}
	\label{fig:fig5}
\end{figure}
\renewcommand\arraystretch{0.5}
\begin{figure}[t]
	\centering
	\setlength{\tabcolsep}{0.03cm}
	\begin{tabular}{ccccc}
		\includegraphics[width=0.194\linewidth]{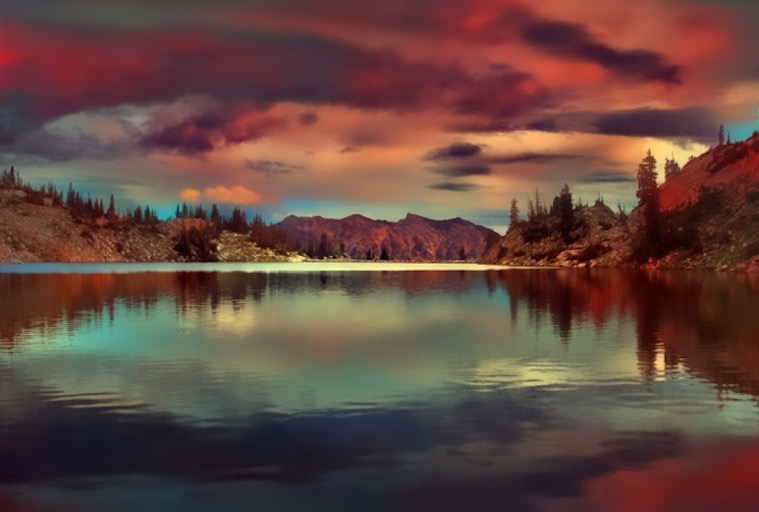}&
		\includegraphics[width=0.194\linewidth]{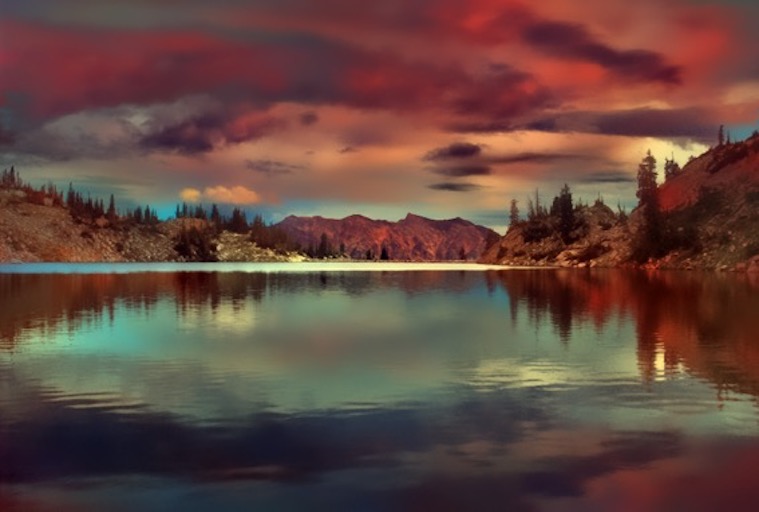}&
		\includegraphics[width=0.194\linewidth]{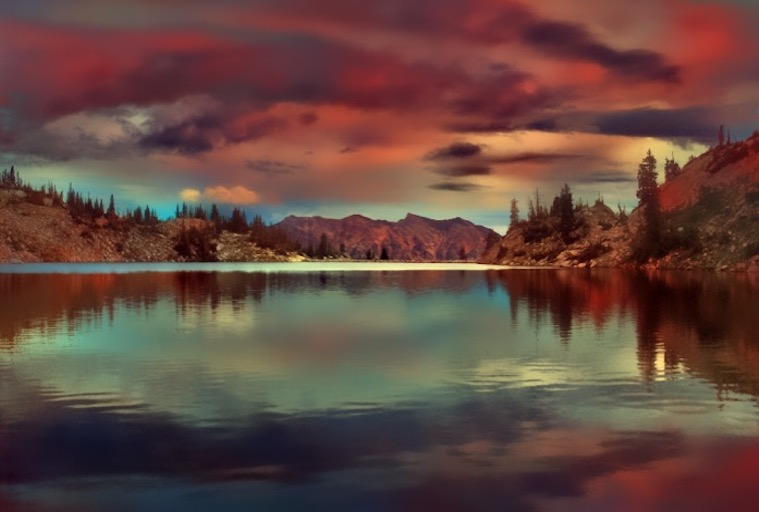}&
		\includegraphics[width=0.194\linewidth]{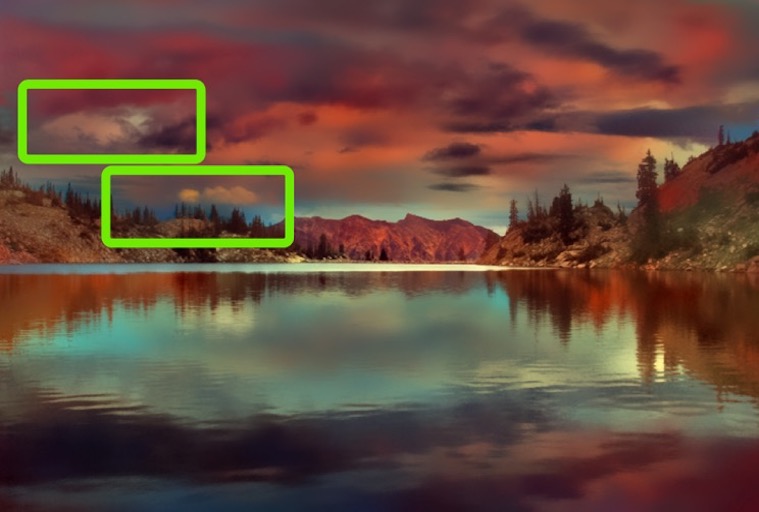}&
		\includegraphics[width=0.194\linewidth]{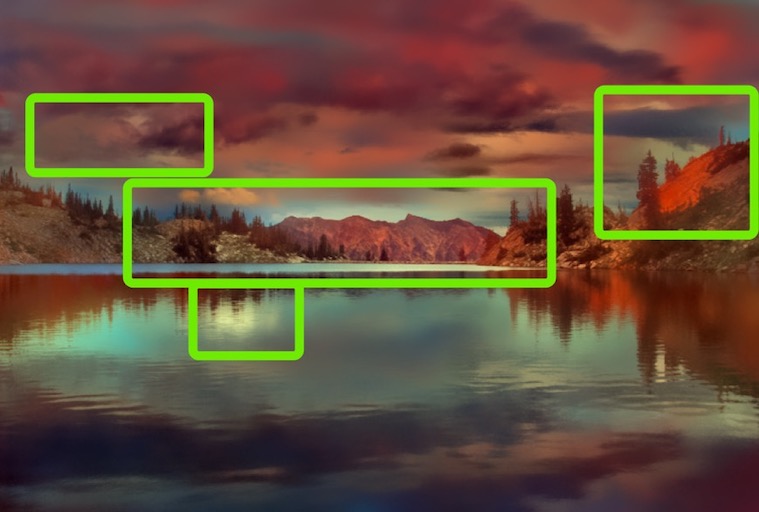}\\
		
		\includegraphics[width=0.194\linewidth]{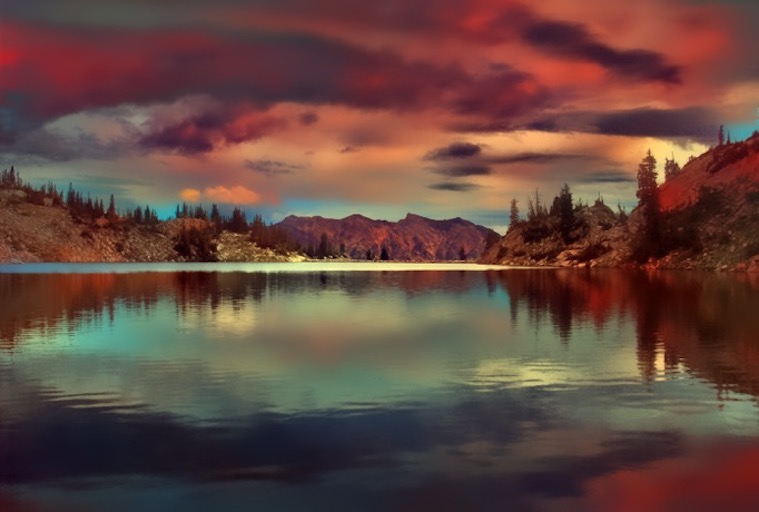}&
		\includegraphics[width=0.194\linewidth]{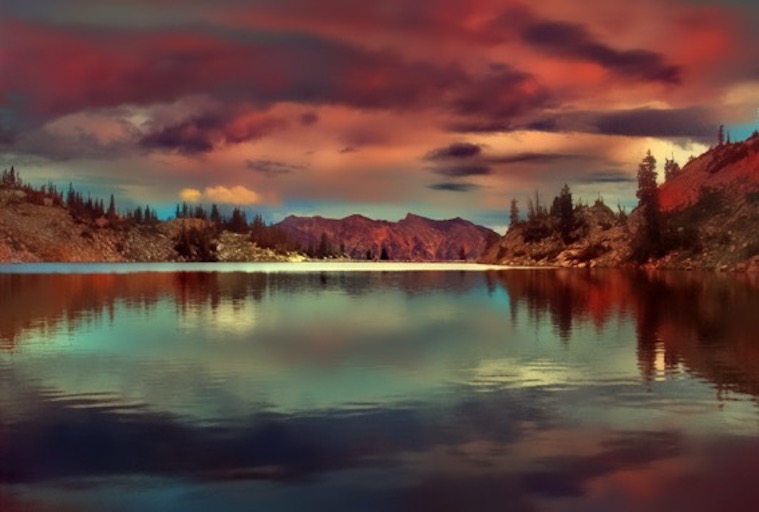}&
		\includegraphics[width=0.194\linewidth]{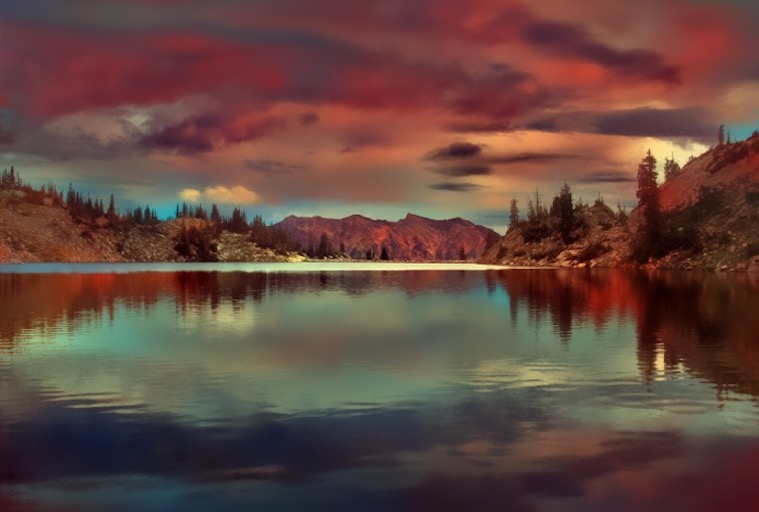}&
		\includegraphics[width=0.194\linewidth]{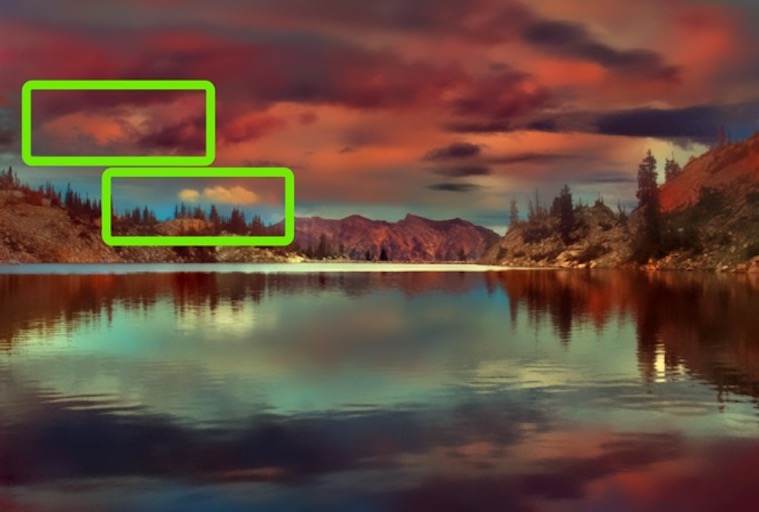}&
		\includegraphics[width=0.194\linewidth]{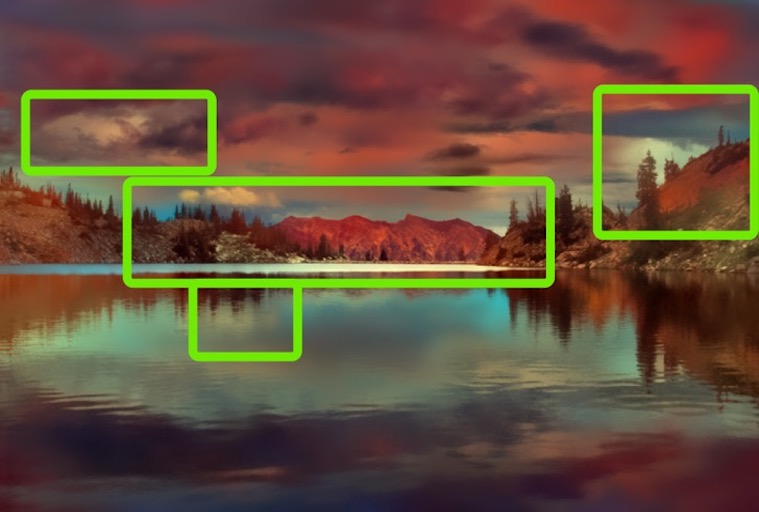}\\
		
		\scriptsize $\lambda=0$&\scriptsize $\lambda=0.4$&\scriptsize $\lambda=0.6$&\scriptsize $\lambda=0.8$&\scriptsize $\lambda=1.0$\\
	\end{tabular}
	\caption{Trade-off between diversity and quality by varying diversity hyperparameter $\lambda$ in method~\cite{li2018closed} (+ our DFP).
	}
	\label{fig:pw}
\end{figure}
\section{Experimental Results}
\label{exp}
\subsection{Implementation Details}
\label{impd}
We incorporate our deep feature perturbation into three existing WCT-based methods which are used for different style transfer tasks, \ie, \cite{li2017universal} for artistic style transfer, \cite{sheng2018avatar} for semantic-level style transfer and \cite{li2018closed} for photo-realistic style transfer. Except for replacing the WCTs with our PWCTs, we do not modify anything else, such as pre-trained models, pre-processing or post-processing operations, etc. If not specifically stated, in all experiments, the stylization weight $\alpha$ of our diversified version is consistent with the original version, and the random noise matrix $N$ is sampled from the standard normal distribution. We fine-tune the diversity hyperparameter $\lambda$ to make our quality similar to previous works, \ie, 0.6 for~\cite{li2017universal}, 0.5 for~\cite{sheng2018avatar} and 1 for~\cite{li2018closed}. We will discuss these settings in the following sections. Our code is available at: \url{https://github.com/EndyWon/Deep-Feature-Perturbation}.

\begin{figure*}[htbp]
	\centering
	\includegraphics[width=1\linewidth]{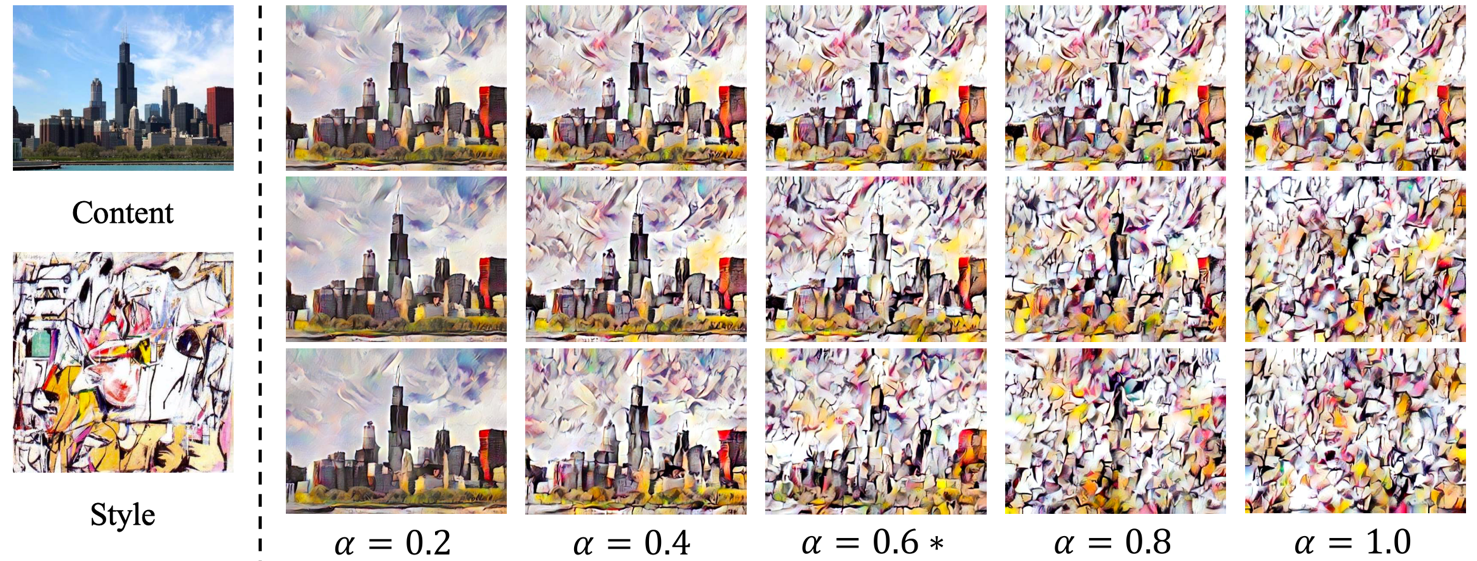}
	\caption{Relation between diversity and stylization strength. Each column (except for the first one) shows the results obtained by different $\alpha$ values (stylization strength). The top row shows the results of the original method~\cite{li2017universal}. The middle row shows the results obtained by setting $\lambda=0.6$ (the default diversity strength) for our diversified version of~\cite{li2017universal}. The bottom row shows the results obtained by increasing the value of $\lambda$ to $1$ for our diversified version of~\cite{li2017universal}. $\alpha=0.6$ is the default stylization setting of~\cite{li2017universal}.
	}
	\label{fig:ss}
\end{figure*}
\renewcommand\arraystretch{0.5}
\begin{figure}[t]
	\centering
	\setlength{\tabcolsep}{0.03cm}
	\begin{tabular}{ccccc}
		\includegraphics[width=0.193\linewidth]{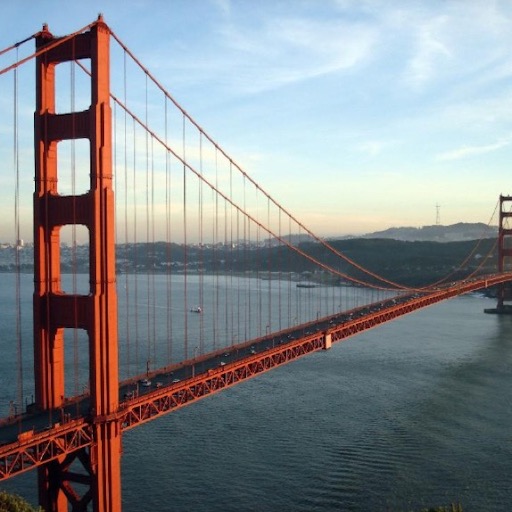}&
		\includegraphics[width=0.193\linewidth]{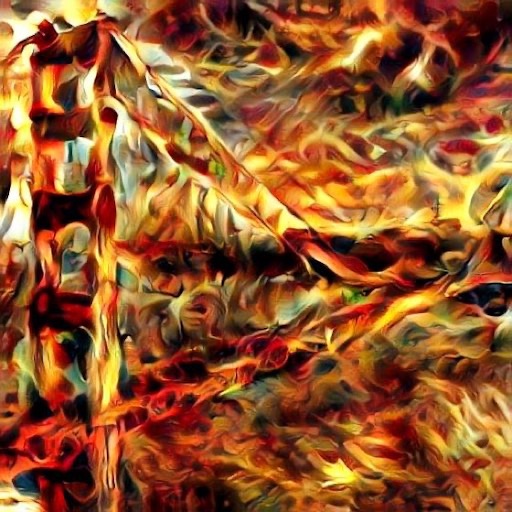}&
		\includegraphics[width=0.193\linewidth]{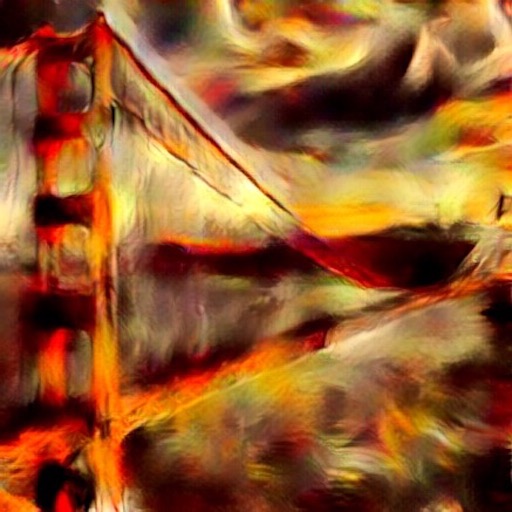}&
		\includegraphics[width=0.193\linewidth]{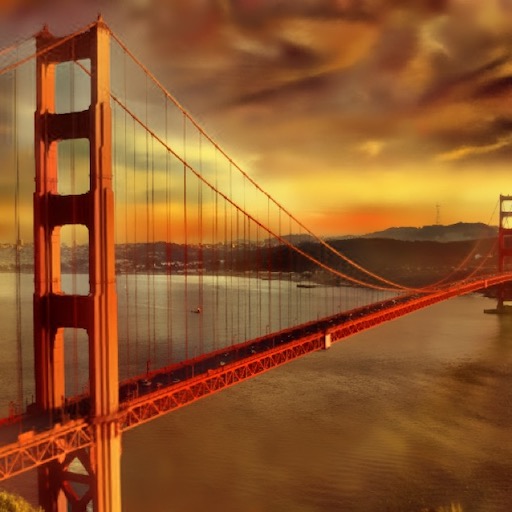}&
		\includegraphics[width=0.193\linewidth]{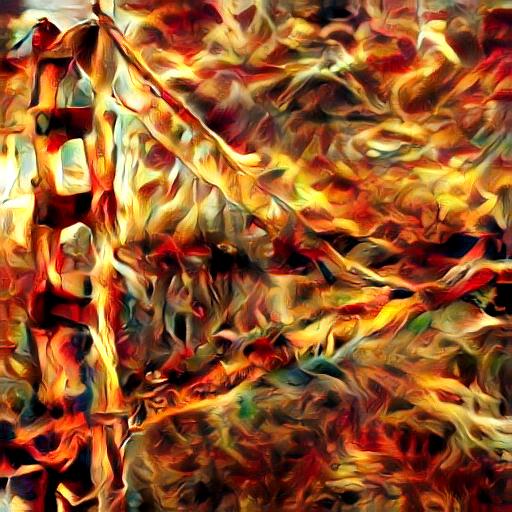}
		\\
		
		\includegraphics[width=0.193\linewidth]{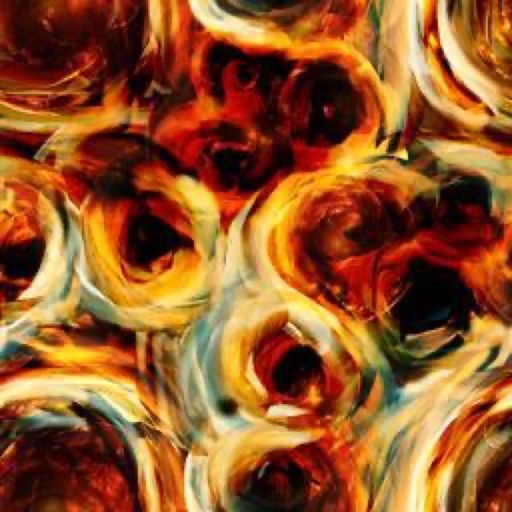}&
		\includegraphics[width=0.193\linewidth]{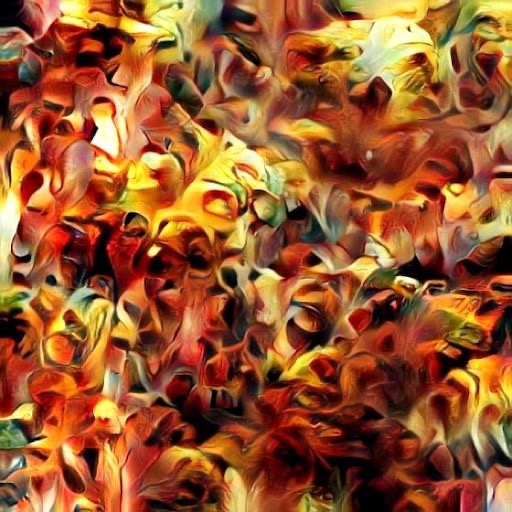}&
		\includegraphics[width=0.193\linewidth]{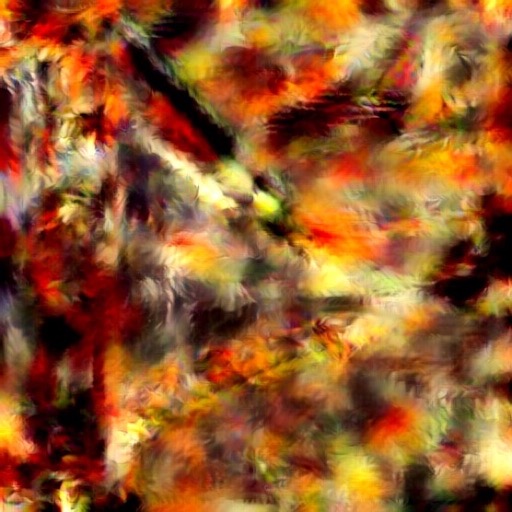}&
		\includegraphics[width=0.193\linewidth]{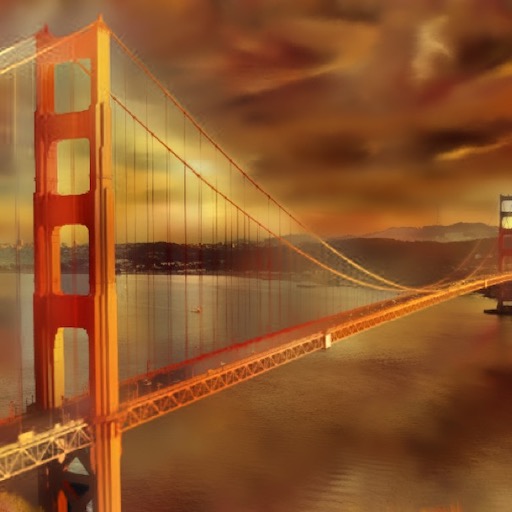}&
		\includegraphics[width=0.193\linewidth]{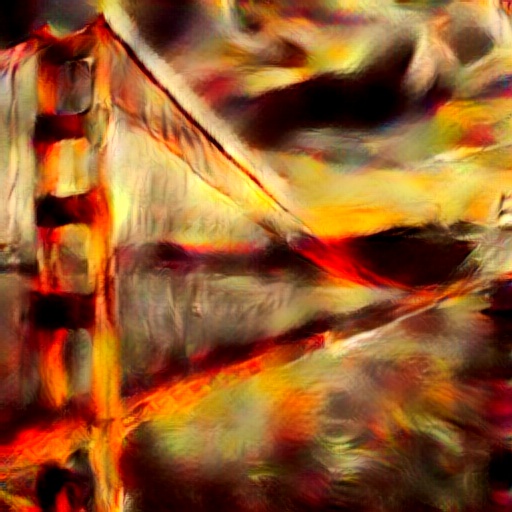}
		\\
		
		\scriptsize Inputs&\scriptsize Li~\etal~\cite{li2017universal}&\scriptsize Sheng~\etal~\cite{sheng2018avatar}&\scriptsize Li~\etal~\cite{li2018closed} &\scriptsize Varied Sampling\\
	\end{tabular}
	\caption{Relation between orthogonal noise matrix and generated result. The first column shows the input content (top) and style (bottom) images. The second to fourth columns show the results obtained by using the orthogonal noise matrix (top) and original random noise matrix (bottom) to perturb the methods~\cite{li2017universal,sheng2018avatar,li2018closed}, respectively. The last column shows the results obtained by varying the sampling distribution of the orthogonal noise matrix for methods~\cite{li2017universal} (top) and~\cite{sheng2018avatar} (bottom).
	}
	\label{fig:fig6}
\end{figure}
\subsection{Ablation Study}
\label{abstudy}
{\bf Single-level Perturbation versus Multi-level Perturbation.} We study the effects of single-level perturbation and multi-level perturbation on two WCT-based methods~\cite{li2017universal,li2018closed}, since they both use the multi-level stylization (while the method~\cite{sheng2018avatar} only uses a single-level stylization). To perturb only specific levels, we set the diversity hyperparameter $\lambda$ of the selected levels to default values (\ie, 0.6 for~\cite{li2017universal} and 1 for~\cite{li2018closed}), and the other levels to 0. As shown in the top row of Fig.~\ref{fig:fig2}, when we perturb separately from the deepest level (I5) to the shallowest level (I1), the quality decreases accordingly. This phenomenon exists in the top row of Fig.~\ref{fig:fig3} as well. We analyze the reason may be that the deeper level stylizes more low-frequency coarse characteristics while the shallower level stylizes more high-frequency fine characteristics, so adding noise into the shallower levels will affect the pixel performance of the final results. Perturbing at the deepest level can achieve comparable stylization quality as the original methods (see I5 in Fig.~\ref{fig:fig2} and I4 in Fig.~\ref{fig:fig3}). On the other hand, multi-level perturbation introduces noise into multiple levels, as shown in the bottom rows of Fig.~\ref{fig:fig2} and Fig.~\ref{fig:fig3}. We can see that introducing too much noise is unnecessary and will reduce the quality of stylization. We also compare the run-time in Table~\ref{tab1}. Note that for method~\cite{li2018closed}, we only consider the stylization time. Compared with the original methods (column 2), the incremental run-time decreases when we perturb the shallower levels. Nevertheless, the deepest-level perturbation only increases a very small amount of time (in {\bf bold}).

\renewcommand\arraystretch{0.7}
\begin{figure*}[h]
	\centering
	\setlength{\tabcolsep}{0.02cm}
	\begin{tabular}{cp{0.1cm}ccccccc}
		\includegraphics[width=0.12\linewidth]{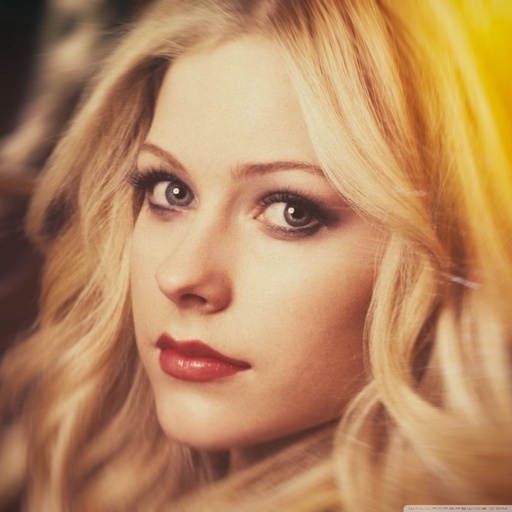}&&
		\includegraphics[width=0.12\linewidth]{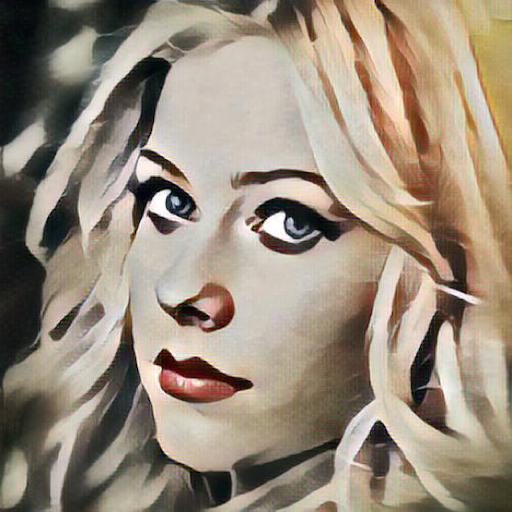}&
		\includegraphics[width=0.12\linewidth]{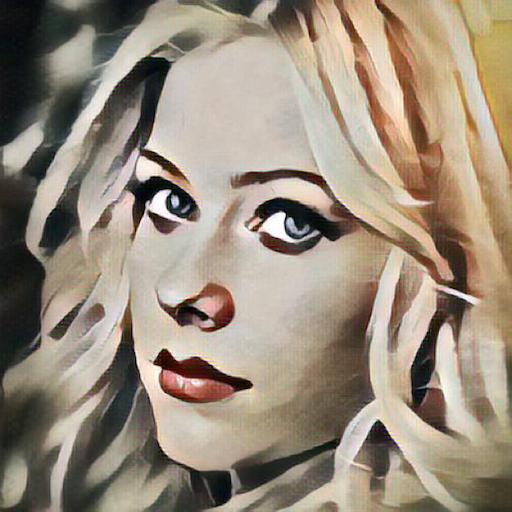}&
		\includegraphics[width=0.12\linewidth]{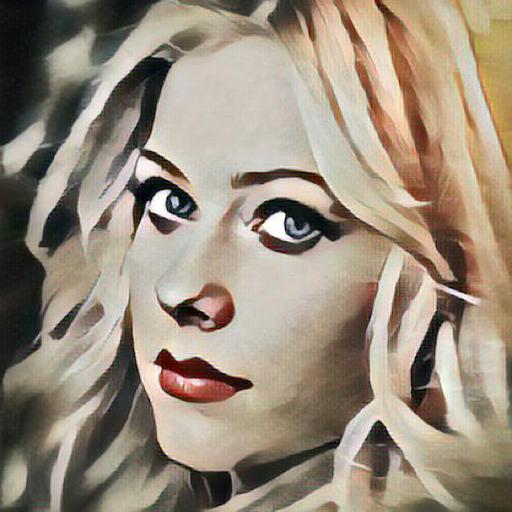}&
		\includegraphics[width=0.12\linewidth]{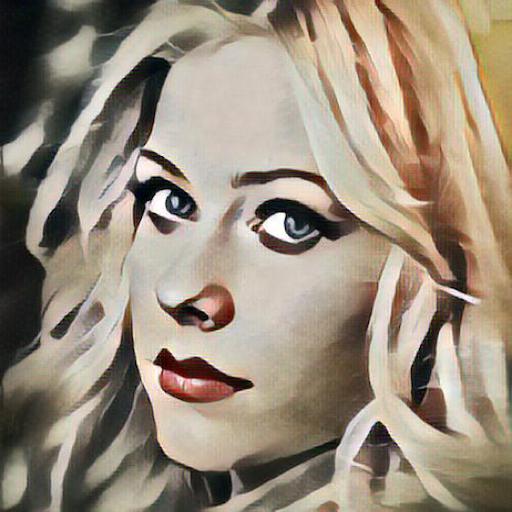}&
		\includegraphics[width=0.12\linewidth]{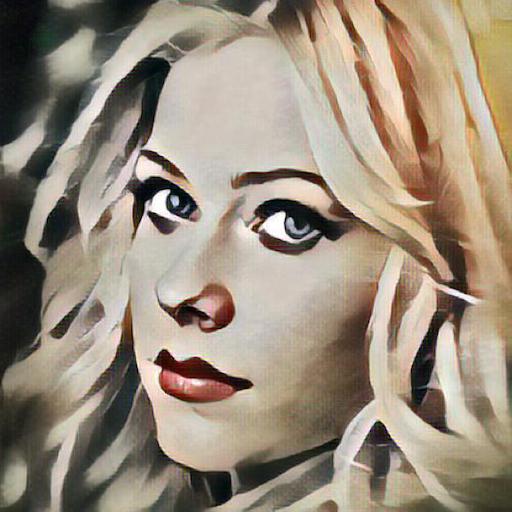}&
		\includegraphics[width=0.12\linewidth]{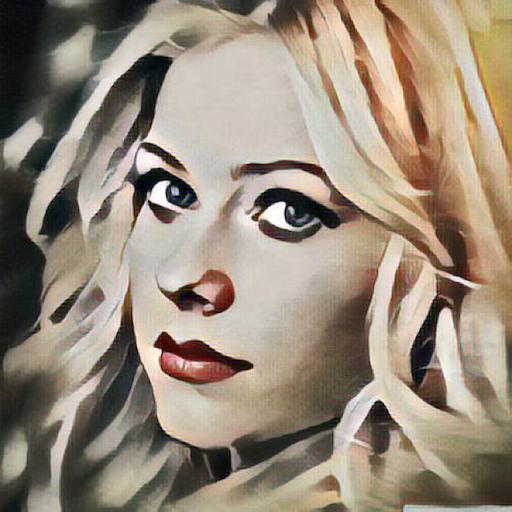}&
		\includegraphics[width=0.12\linewidth]{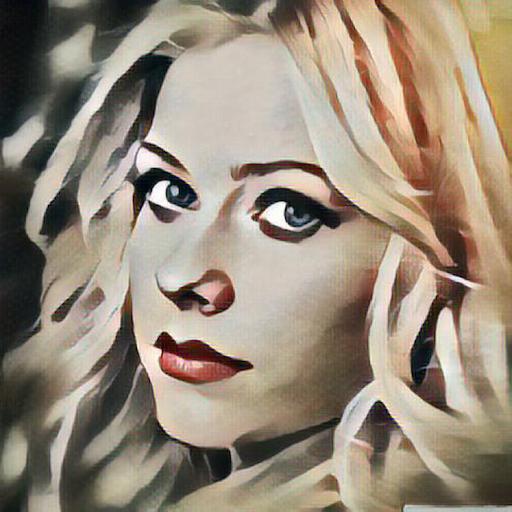}
		\\
		Content&&\multicolumn{7}{|c|}{Li~\etal~\cite{li2017diversified}}\\
		
		\includegraphics[width=0.12\linewidth]{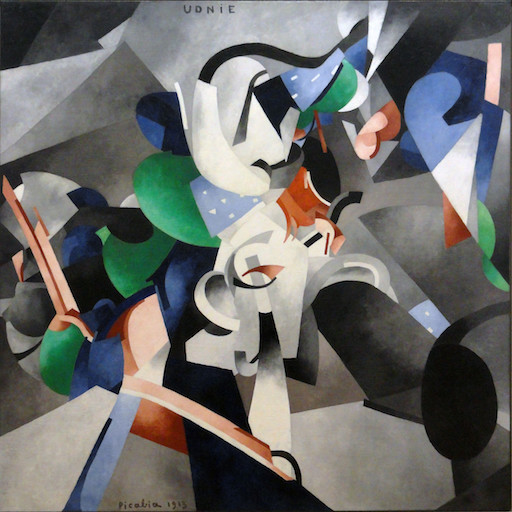}&&
		\includegraphics[width=0.12\linewidth]{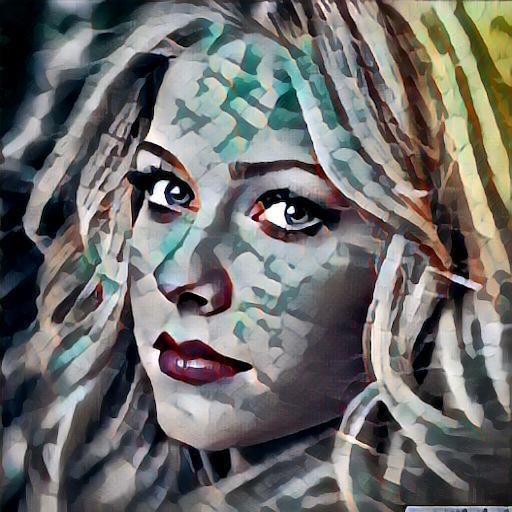}&
		\includegraphics[width=0.12\linewidth]{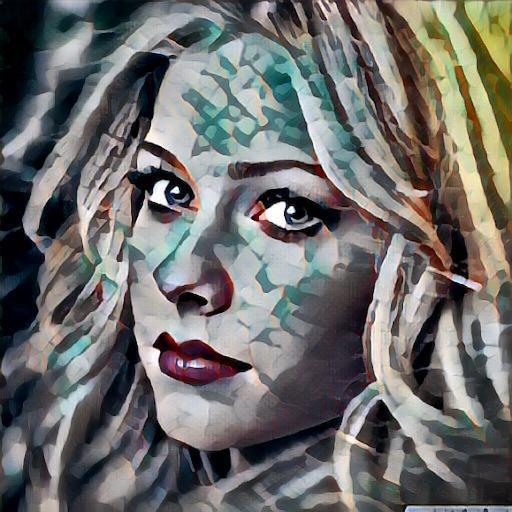}&
		\includegraphics[width=0.12\linewidth]{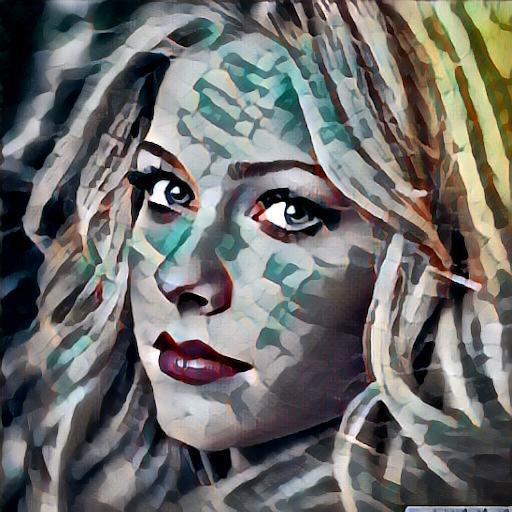}&
		\includegraphics[width=0.12\linewidth]{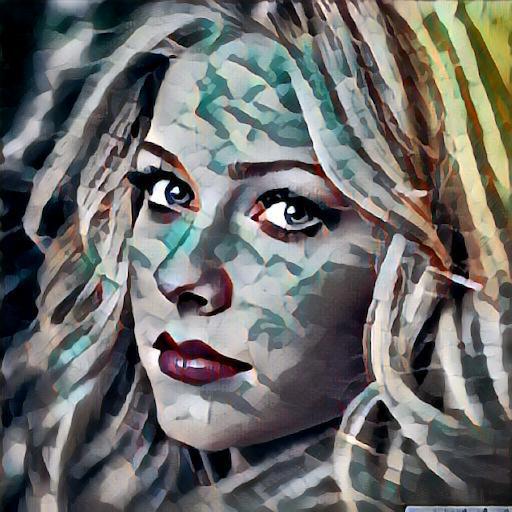}&
		\includegraphics[width=0.12\linewidth]{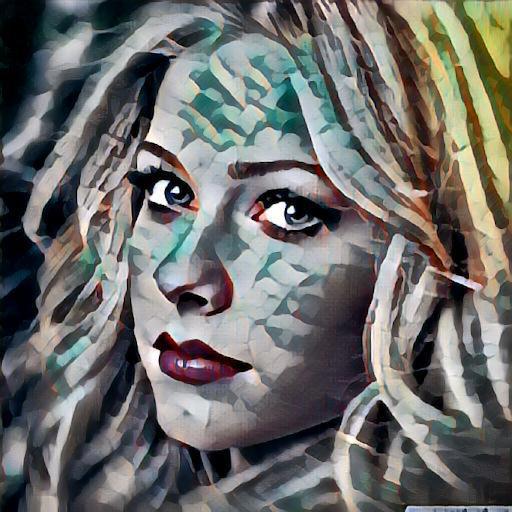}&
		\includegraphics[width=0.12\linewidth]{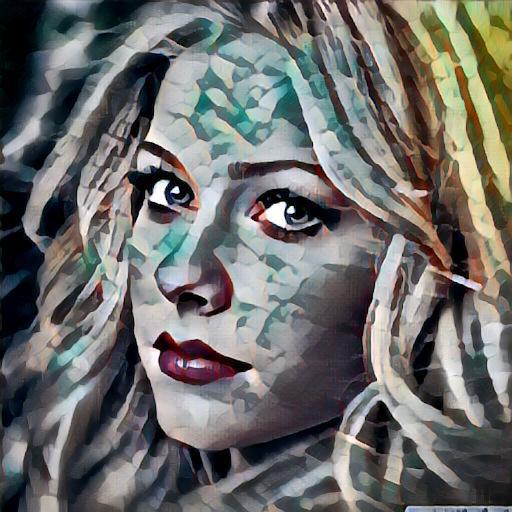}&
		\includegraphics[width=0.12\linewidth]{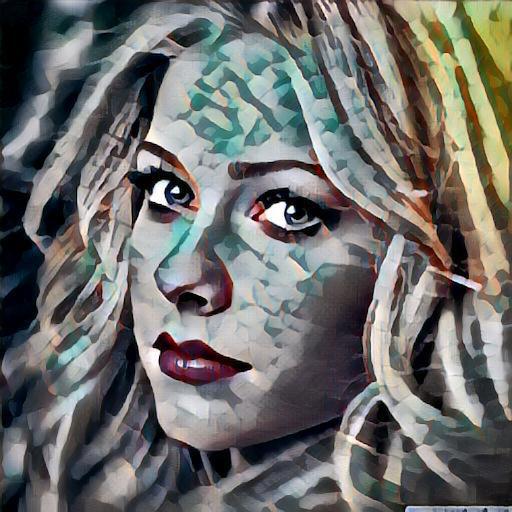}
		\\
		Style&&\multicolumn{7}{|c|}{Ulyanov~\etal~\cite{ulyanov2017improved}}\\
		
		\includegraphics[width=0.12\linewidth]{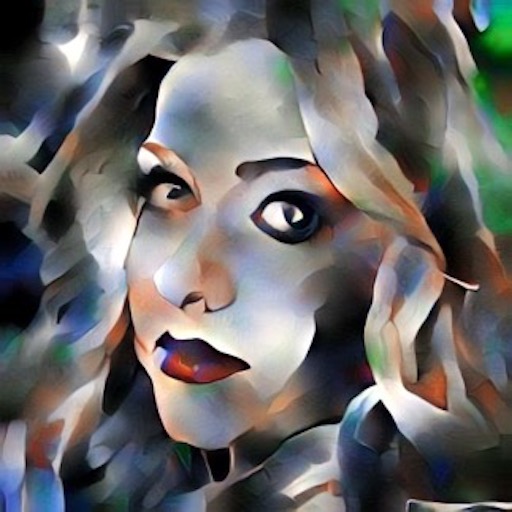}&&
		\includegraphics[width=0.12\linewidth]{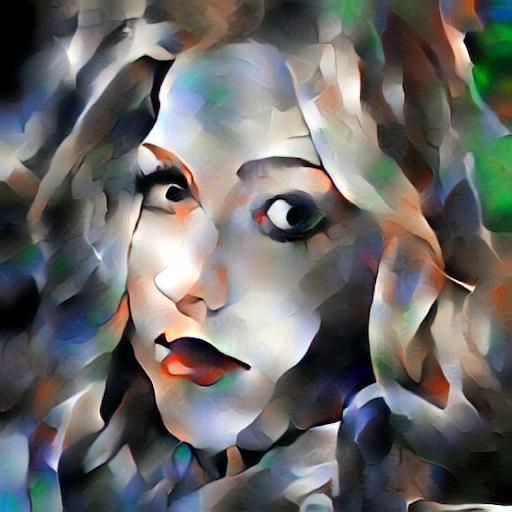}&
		\includegraphics[width=0.12\linewidth]{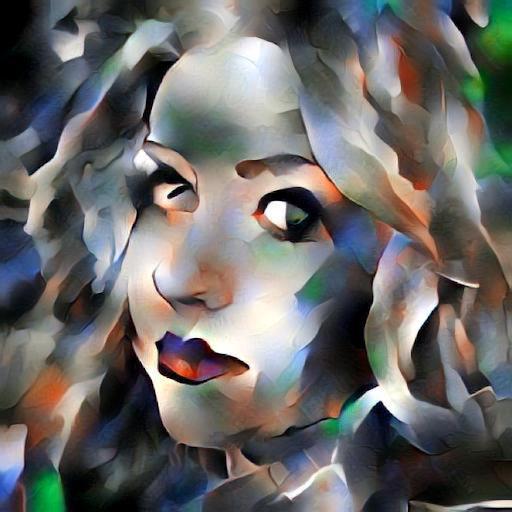}&
		\includegraphics[width=0.12\linewidth]{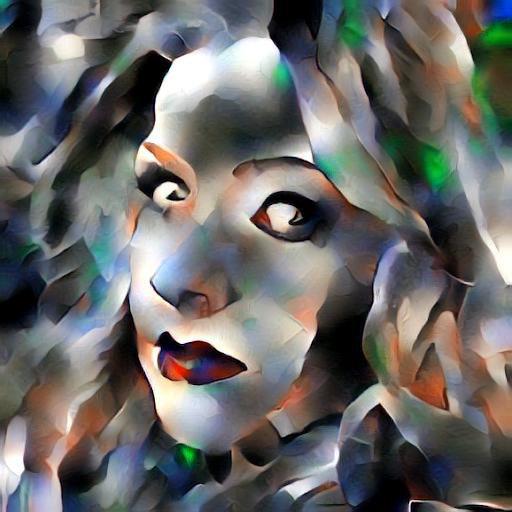}&
		\includegraphics[width=0.12\linewidth]{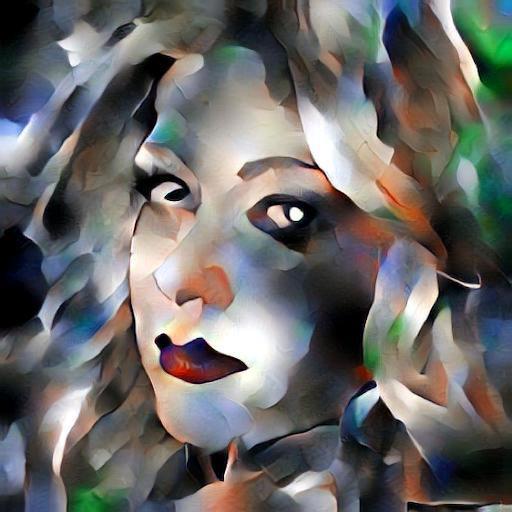}&
		\includegraphics[width=0.12\linewidth]{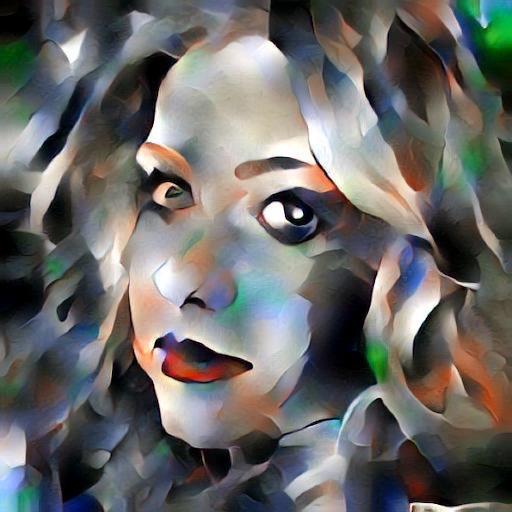}&
		\includegraphics[width=0.12\linewidth]{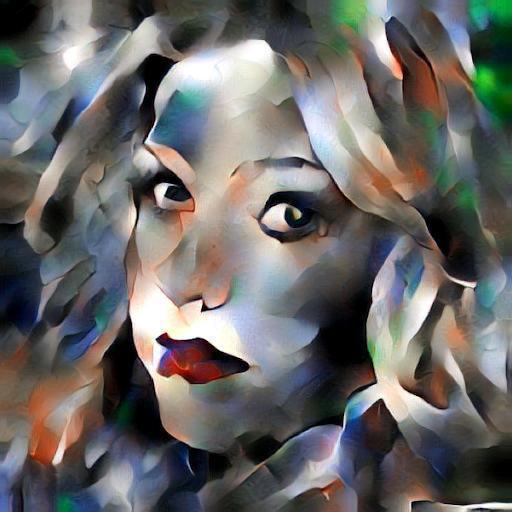}&
		\includegraphics[width=0.12\linewidth]{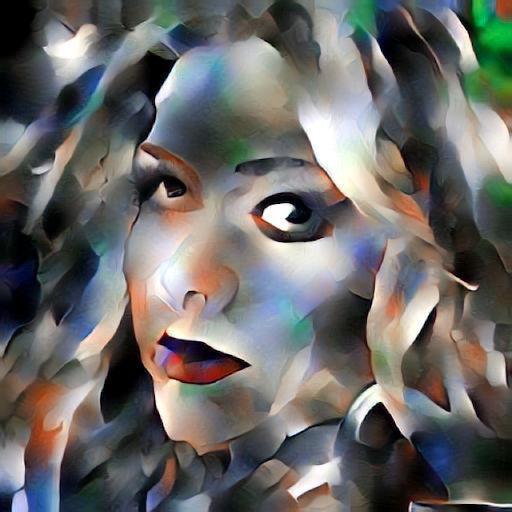}
		\\
		Li~\etal~\cite{li2017universal}&&\multicolumn{7}{|c|}{Li~\etal~\cite{li2017universal} + our DFP}\\
		
		\includegraphics[width=0.12\linewidth]{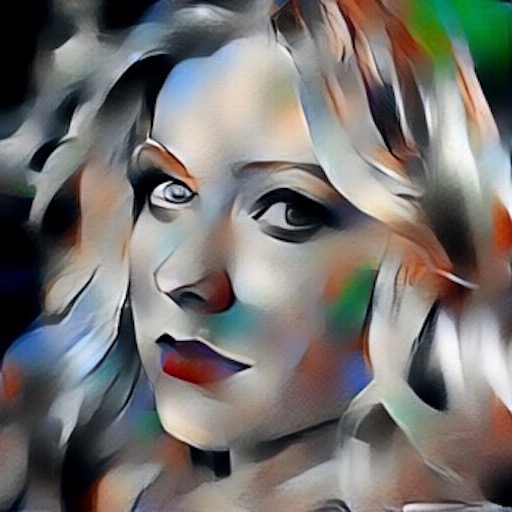}&&
		\includegraphics[width=0.12\linewidth]{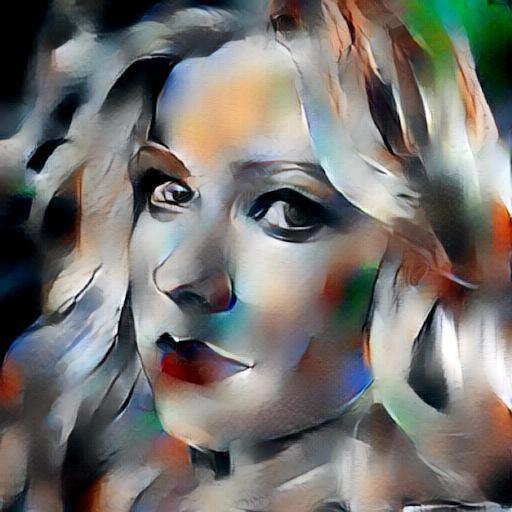}&
		\includegraphics[width=0.12\linewidth]{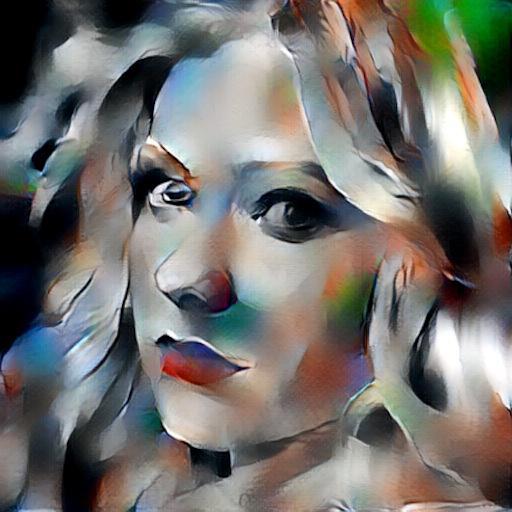}&
		\includegraphics[width=0.12\linewidth]{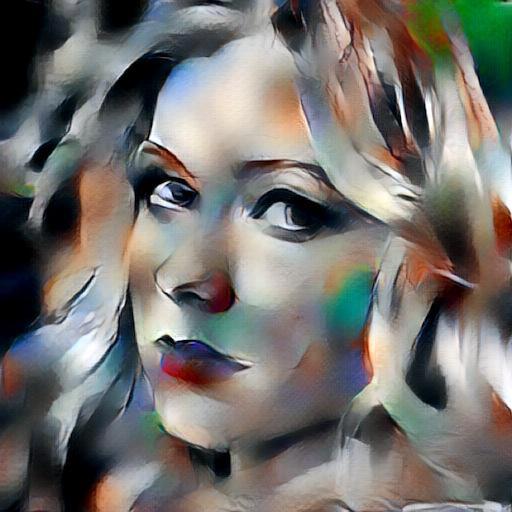}&
		\includegraphics[width=0.12\linewidth]{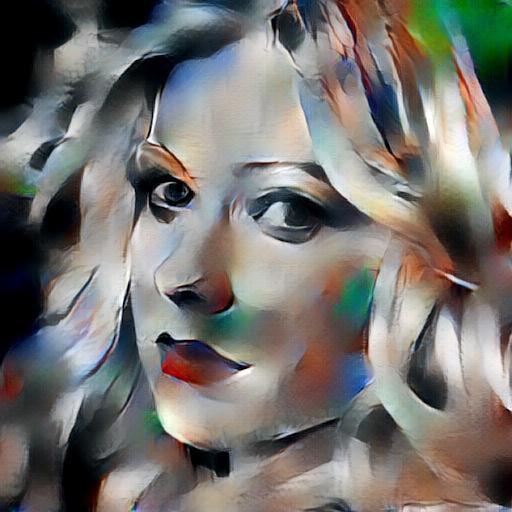}&
		\includegraphics[width=0.12\linewidth]{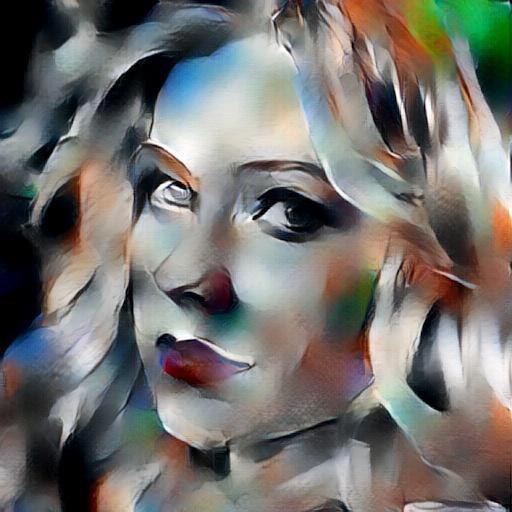}&
		\includegraphics[width=0.12\linewidth]{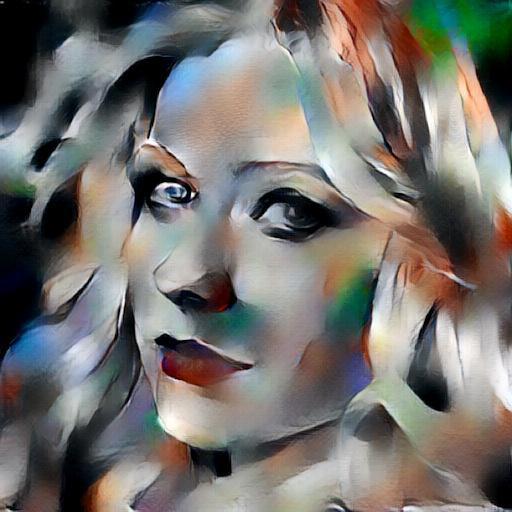}&
		\includegraphics[width=0.12\linewidth]{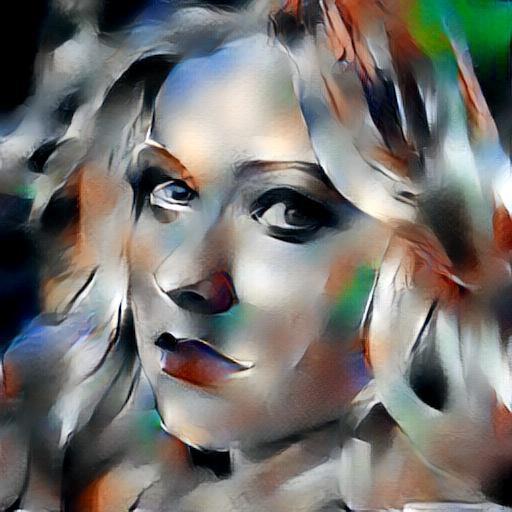}
		\\
		Sheng~\etal~\cite{sheng2018avatar}&&\multicolumn{7}{|c|}{Sheng~\etal~\cite{sheng2018avatar} + our DFP}\\
		
		\includegraphics[width=0.12\linewidth]{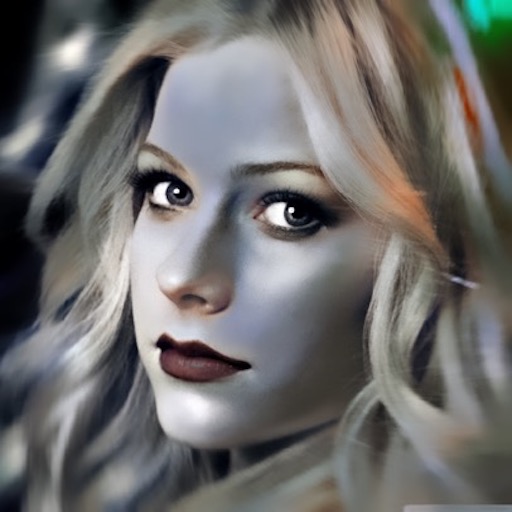}&&
		\includegraphics[width=0.12\linewidth]{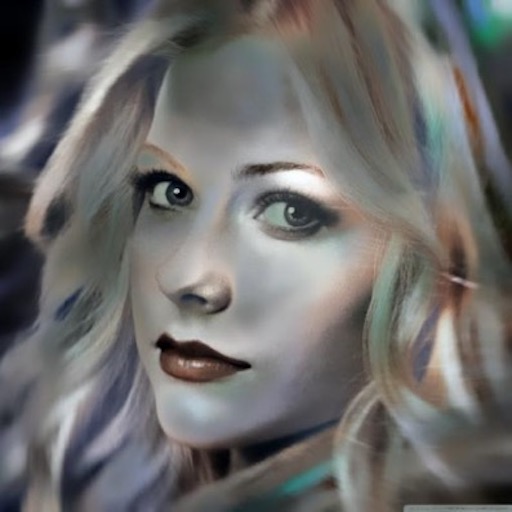}&
		\includegraphics[width=0.12\linewidth]{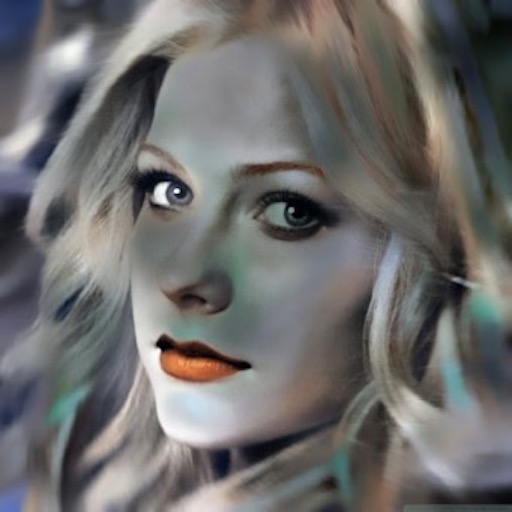}&
		\includegraphics[width=0.12\linewidth]{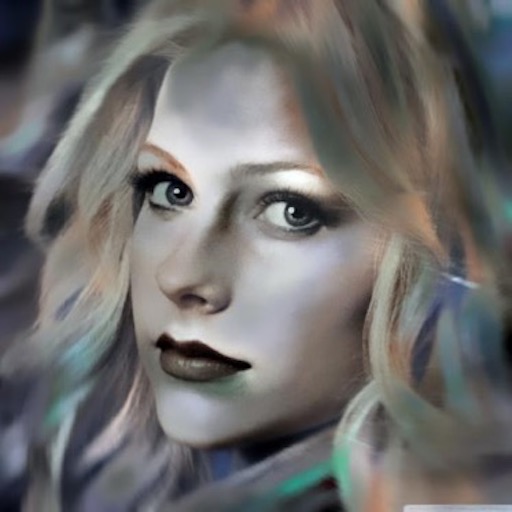}&
		\includegraphics[width=0.12\linewidth]{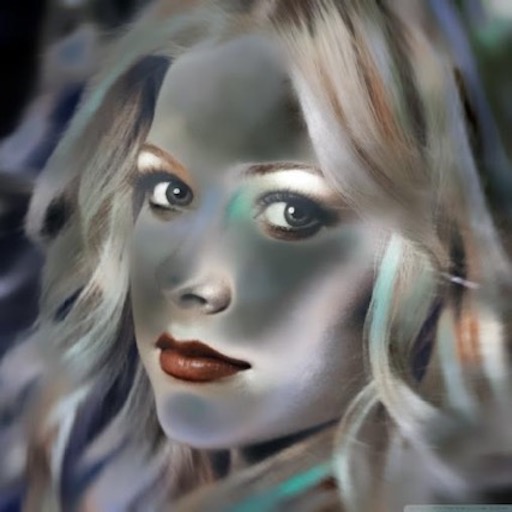}&
		\includegraphics[width=0.12\linewidth]{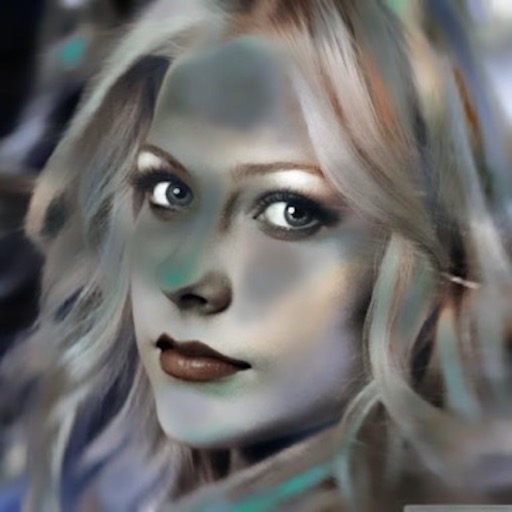}&
		\includegraphics[width=0.12\linewidth]{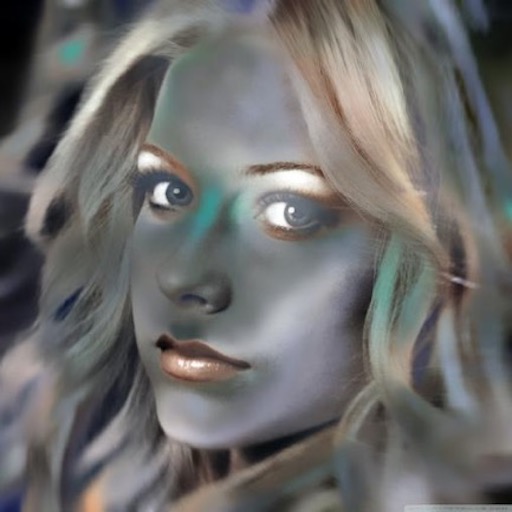}&
		\includegraphics[width=0.12\linewidth]{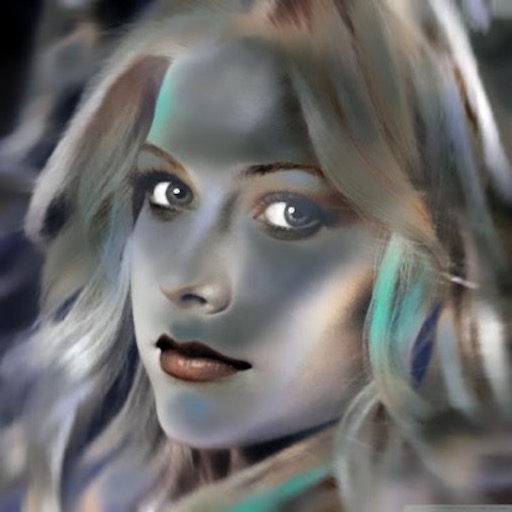}
		\\
		Li~\etal~\cite{li2018closed}&&\multicolumn{7}{|c|}{Li~\etal~\cite{li2018closed} + our DFP}\\

	\end{tabular}
	
	\caption{Qualitative comparisons of different diversified style transfer methods. The first column (from top to bottom) shows inputs and original outputs of~\cite{li2017universal,sheng2018avatar,li2018closed}. The other columns (from top to bottom) show diverse outputs of~\cite{li2017diversified,ulyanov2017improved} and~\cite{li2017universal,sheng2018avatar,li2018closed} (+ our DFP).
	}
	\label{fig:fig7}
\end{figure*}
{\bf Trade-off between Diversity and Quality.} In Eq.~(\ref{eq8}), we introduce a diversity hyperparameter $\lambda$ to provide user controls on the trade-off between diversity and quality. Different methods may require different $\lambda$ values. In this part, we demonstrate the impact of different $\lambda$ values on methods~\cite{li2017universal,sheng2018avatar,li2018closed} while keeping their default stylization settings. For method~\cite{li2017universal} and~\cite{li2018closed}, we only perturb the deepest level as suggested in the former sections. For method~\cite{sheng2018avatar}, we perturb its bottleneck layer as it only uses a single-level stylization. The results are shown in Fig.~\ref{fig:fig4},~\ref{fig:fig5} and~\ref{fig:pw}. As we can see, the degree of diversity rises with the increase of $\lambda$ values, but for method~\cite{li2017universal} and~\cite{sheng2018avatar} (Fig.~\ref{fig:fig4} and~\ref{fig:fig5}), the quality is obviously reduced when large $\lambda$ values are applied. However, this problem does not arise in method~\cite{li2018closed} (Fig.~\ref{fig:pw}), it may be because this method~\cite{li2018closed} contains a smoothing step to remove noticeable artifacts and it suppresses the emergence of diversity to some extent, which will also be verified by the quantitative comparisons in later Section~\ref{com}. For trade-offs, we finally adopt 0.6, 0.5 and 1 for the default $\lambda$ values of~\cite{li2017universal},~\cite{sheng2018avatar} and~\cite{li2018closed}, respectively.

{\bf Relation between Diversity and Stylization Strength.} The diversity is also related to the stylization strength. Taking method~\cite{li2017universal} as an example, Fig.~\ref{fig:ss} demonstrates the relation between these two aspects. Comparing the top two rows, we can observe that for our default diversity setting ($\lambda=0.6$), it works well for the situations where the stylization strength $\alpha\leq 0.6$, but destroys the content structure for those with larger $\alpha$ values. We set a larger diversity strength ($\lambda=1$) in the bottom row, and we can observe that it still works fine for those with lower stylization strength (\eg, $\alpha\leq 0.4$). That is to say, we can set a larger diversity strength for a smaller stylization strength. In fact, as we have analyzed in Section~\ref{DFP}, our diversity may affect the content information from $\hat{F_c}$ (Eq. (\ref{eq5})), so the content structure will be overwhelmed by the style patterns when the value of $\lambda$ is too high, as validated in the last two columns. Therefore, the tradeoff between stylization strength ($\alpha$) and diversity strength ($\lambda$) should be considered. Nevertheless, in practice, users only need to first determine the optimal stylization strength $\alpha$ (usually the default one) for different methods, and then adjust the appropriate $\lambda$ values to keep the quality. Besides, in each method, our results have verified that the constant $\lambda$ value can work {\em stably} on different content and style inputs.

{\bf Locations to Insert the Orthogonal Noise Matrix.} In Section~\ref{DFP}, we have mentioned three places to insert the orthogonal noise matrix in Eq.~(\ref{eq6}), \ie, between $D_s^{\frac{1}{2}}$ and $E_s^T$, between $E_s^T$ and $\hat{F_c}$, and on the right side of $\hat{F_c}$. We conduct the same experiments for each of them and find that there is no difference in qualitative comparisons. But in quantitative comparisons, \eg, run-time and computation requirements, there are some differences. This is mainly due to the different computation of matrix multiplication caused by the different size of noise matrix. As we have analyzed in Section~\ref{DFP}, when we insert the orthogonal noise matrix ${\bf Z}$ between $D_s^{\frac{1}{2}}$ and $E_s^T$, the size of ${\bf Z}$ is only $(C-k)\times (C-k)$, where $C$ is the number of channels and $k$ is the number of small singular values in $D_s^{\frac{1}{2}}$. For the other two cases, since the shapes of $E_s^T$ and $\hat{F_c}$ are $(C-k)\times C$ and $C\times H_cW_c$, respectively (where $H_c$, $W_c$ are the height and width of the content feature), the size of ${\bf Z}$ should be $C\times C$ if we insert it between $E_s^T$ and $\hat{F_c}$, and $H_cW_c\times H_cW_c$ if we insert it on the right side of $\hat{F_c}$. Generally, for the deepest level, $C-k<C<H_cW_c$, so we eventually insert ${\bf Z}$ between $D_s^{\frac{1}{2}}$ and $E_s^T$ as this may consume the least computation and run-time.

\newcommand{\tabincell}[2]{\begin{tabular}{@{}#1@{}}#2\end{tabular}}
\begin{table}[h]
	\caption{Quantitative comparisons of different methods. We measure diversity using average Pixel distance and LPIPS distance~\cite{zhang2018unreasonable}.}
	\centering
	\hspace{1.5cm}
	\begin{tabular}{ccc}
		\toprule
		Method&\tabincell{c}{Pixel \\Distance}&\tabincell{c}{LPIPS \\Distance}\\
		\midrule
		Li \etal~\cite{li2017diversified}&0.080&0.175\\
		Ulyanov \etal~\cite{ulyanov2017improved}&0.077&0.163\\
		Li \etal~\cite{li2017universal}&0.000&0.000\\
		Sheng \etal~\cite{sheng2018avatar}&0.000&0.000\\
		Li \etal~\cite{li2018closed}&0.000&0.000\\
		\midrule
		Li \etal~\cite{li2017universal} + our DFP&{\bf 0.162}&{\bf 0.431}\\
		Sheng \etal~\cite{sheng2018avatar} + our DFP&{\bf 0.102}&{\bf 0.264}\\
		Li \etal~\cite{li2018closed} + our DFP&{\bf 0.091}&{\bf 0.203}\\
		\bottomrule
	\end{tabular}
	\label{tab2}
\end{table}
{\bf Relation between Orthogonal Noise Matrix and Generated Result.} To verify the importance and necessity of the orthogonal noise matrix ${\bf Z}$ in our DFP, we compare it with the original random noise matrix $N$, and also discuss the influence of its sampling distribution. The results are shown in Fig.~\ref{fig:fig6}, as we can see, using the original random noise matrix produces low quality results (see column 2 to 4 in bottom row). The results obtained by~\cite{li2017universal} and~\cite{sheng2018avatar} are just like combinations of texture and noise, which drown out the content information. Compared with the former two,~\cite{li2018closed} can maintain the content information as much as possible even with the original random noise perturbation. This may be because it consists of two steps, and the second step removes noticeable artifacts to maintain the structure of the content image. But as the result shows, the quality is still significantly reduced. Similar to the former experiments, we also adjust the values of $\alpha$ and $\lambda$ for original random noise perturbation, but the poor generation effect still cannot be alleviated. To explore the influence of sampling distribution of orthogonal noise matrix, we use uniform distribution instead of the standard normal distribution for method~\cite{li2017universal} (see the last column in top row), and vary the mean and standard deviation of normal distribution for method~\cite{sheng2018avatar} (see the last column in bottom row). As we can see, the generated images do not show a significant difference from the default ones, which indicates that the key factor affecting the result is the orthogonality of noise ${\bf Z}$, rather than its sampling distribution.

\subsection{Comparisons}
\label{com}
In this section, we incorporate our DFP into methods~\cite{li2017universal,sheng2018avatar,li2018closed} and compare them with other diversified style transfer methods~\cite{li2017diversified,ulyanov2017improved} from both qualitative and quantitative aspects. For methods~\cite{li2017diversified} and~\cite{ulyanov2017improved}, we run the author-released codes or pre-trained models with the default configurations. For our methods, we use the default settings as described in Section~\ref{impd}.

{\bf Qualitative Comparisons.} We show qualitative comparison results in Fig.~\ref{fig:fig7}. We observe that~\cite{li2017diversified} and~\cite{ulyanov2017improved} only produce subtle diversity (\eg, slight changes in the faces), which does not contain any meaningful variation. By contrast, for the methods with our DFP, the results show a distinct diversity (\eg, the faces, the hairs, the backgrounds, and even the eyes). Compared with the original outputs, the results obtained by incorporating our DFP are almost without quality degradation. 

{\bf Quantitative Comparisons.} We compute the average distance of sample pairs in pixel space and deep feature space to measure the diversity, respectively. For each method, we use 6 content images and 6 style images to get 36 different combinations, and for each combination, we obtain 20 outputs. There are totally 6840 pairs (each pair has the same content and style) of outputs generated by each method, we compute the average distance between them.

In pixel space, we directly compute the average pixel distance in RGB channels, which can be formulated as follows:
\begin{equation}
d_{pixel} (\vec{x}_1, \vec{x}_2)= \frac{||\vec{x}_1-\vec{x}_2||_1}{W\times H\times 255\times 3},
\label{pd}
\end{equation}
where $\vec{x}_1$ and $\vec{x}_2$ denote the two images to compute the pixel distance. $W$ and $H$ are their width and height (they should have the same resolution).

In deep feature space, we use the LPIPS (Learned Perceptual Image Patch Similarity) metric proposed by Zhang~\etal~\cite{zhang2018unreasonable}. It computes distance in AlexNet~\cite{krizhevsky2014one} feature space ($conv1\_5$, pre-trained on Imagenet~\cite{russakovsky2015imagenet}), with linear weights to better match human perceptual judgments.

As shown in Table~\ref{tab2}, \cite{li2017diversified} and~\cite{ulyanov2017improved} produce low diversity scores in both Pixel and LPIPS distance. Without our DFP, the original methods~\cite{li2017universal,sheng2018avatar,li2018closed} cannot generate diverse results. By incorporating DFP, these methods show great diversity improvement. Note that since the method~\cite{sheng2018avatar} (+ our DFP) is still restricted by some semantic constraints when transferring styles, and method~\cite{li2018closed} (+ our DFP) contains a smoothing step to remove detailed effects, their diversity scores are lower than those of method~\cite{li2017universal} (+ our DFP).

\section{Conclusion}
In this work, we introduce deep feature perturbation (DFP) into the whitening and coloring transform (WCT) to achieve diversified arbitrary style transfer. By incorporating our method, many existing WCT-based methods can be empowered to generate diverse results. Experimental results demonstrate that our approach can greatly increase the diversity while maintaining the quality of stylization. At this stage, we only explore the WCT-based methods, but this learning-free and universal paradigm may inspire a series of more ingenious and effective works in the future. Besides, WCT has also been widely used in many other fields, such as {\em image-to-image translation}~\cite{cho2019image}, {\em GANs}~\cite{siarohin2018whitening}, etc. Therefore, we believe our method may also provide a good inspiration for these research fields.

{\bf Acknowledgments.} We sincerely thank the anonymous reviewers for helping us to improve this paper. This work was supported in part by the Zhejiang science and technology program (No: 2019C03137), and Zhejiang Fund Project (No: LGF18F020006, LY19F020049).

{\small
\bibliographystyle{ieee_fullname}
\bibliography{egbib}
}

\end{document}